\documentclass{article}

\PassOptionsToPackage{numbers,compress}{natbib}

\usepackage[main, final]{neurips_2026}

\usepackage[utf8]{inputenc}
\usepackage[T1]{fontenc}
\usepackage{hyperref}
\usepackage{url}
\usepackage{booktabs}
\usepackage{amsfonts}
\usepackage{amsmath,amssymb,amsthm,mathtools,bm}
\usepackage{nicefrac}
\usepackage{microtype}
\usepackage{xcolor}
\usepackage{graphicx}
\usepackage{multirow}
\usepackage{enumitem}
\usepackage{algorithm}
\usepackage{algpseudocode}
\usepackage{cleveref}
\usepackage{bbm}
\usepackage{makecell}
\usepackage{tabularx}

\hypersetup{colorlinks=true,citecolor=blue,linkcolor=blue,urlcolor=blue}
\setlist[itemize]{leftmargin=*,topsep=2pt,itemsep=1pt,parsep=0pt}
\setlist[enumerate]{leftmargin=*,topsep=2pt,itemsep=1pt,parsep=0pt}

\newtheorem{theorem}{Theorem}
\newtheorem{proposition}{Proposition}
\newtheorem{lemma}{Lemma}
\newtheorem{corollary}{Corollary}
\newtheorem{definition}{Definition}

\newtheorem*{theorem*}{Theorem}
\newtheorem*{proposition*}{Proposition}
\newtheorem*{lemma*}{Lemma}
\newtheorem*{corollary*}{Corollary}
\newtheorem*{fact*}{Fact}

\crefname{assumption}{Assumption}{Assumptions}
\Crefname{assumption}{Assumption}{Assumptions}
\crefname{theorem}{Theorem}{Theorems}
\Crefname{theorem}{Theorem}{Theorems}
\crefname{proposition}{Proposition}{Propositions}
\Crefname{proposition}{Proposition}{Propositions}
\crefname{lemma}{Lemma}{Lemmas}
\Crefname{lemma}{Lemma}{Lemmas}
\crefname{corollary}{Corollary}{Corollaries}
\Crefname{corollary}{Corollary}{Corollaries}
\crefname{definition}{Definition}{Definitions}
\Crefname{definition}{Definition}{Definitions}
\crefname{remark}{Remark}{Remarks}
\Crefname{remark}{Remark}{Remarks}
\crefname{example}{Example}{Examples}
\Crefname{example}{Example}{Examples}
\crefname{fact}{Fact}{Facts}
\Crefname{fact}{Fact}{Facts}

\newcommand{\E}{\mathbb{E}}

\newcommand{\tr}{\mathrm{tr}}
\newcommand{\rank}{\mathrm{rank}}

\newcommand{\SubOpt}{\mathrm{SubOpt}}

\newcommand{\Cqry}{\mathcal{C}_{\mathrm{qry}}}

\newcommand{\methodname}{SURE}
\newcommand{\SURE}{\textsc{\methodname}}
\newcommand{\SUREscore}{\mathrm{\methodname}}

\title{Uncertainty-Guided Reward Labeling for Reinforcement Learning under Limited Feedback}

\author{%
  Renhao Zhang\\
  University of Massachusetts \\
  \texttt{renhaozhang@cs.umass.edu} \\
  \And
  Shreyas Chaudhari\\
  University of Massachusetts \\
  \texttt{schaudhari@cs.umass.edu}\\
  \And
  Bruno Castro da Silva\\
  University of Massachusetts \\
  \texttt{bsilva@cs.umass.edu}
}

\begin{document}
\maketitle

\begin{abstract}
In reinforcement learning from limited feedback (RLLF), only a small fraction of an offline dataset can be labeled with rewards, and the central question is \textit{which} samples should be labeled to learn a high-performing policy from the resulting partially labeled dataset. Prior work formalized this as a reward-selection problem by focusing on the selection stage while treating downstream policy learning as a black box, in a regime where queried rewards are not retained for reward-model training.
We instead study the retained-label setting, where queried rewards can be stored and used to fit a reward model before policy learning.
We bound the suboptimality of the learned policy in terms of two sources of error: one arising from offline RL on a fixed, offline dataset, and the other from uncertainty in the reward model. 
Since reward selection cannot change the offline dataset, the limited labeling budget must be used to strategically reduce reward uncertainty.
Motivated by RLLF's observation that useful rewards tend to keep the agent on high-return trajectories, we introduce \textit{Successor-Guided Uncertainty REduction} (SURE), a novel method that uses successor features to select rewards that are both uncertain and associated with states from which high-valued states are reachable. We formally derive SURE from a bound-induced design objective and characterize its exact one-step marginal gain. Empirically, SURE reaches near full-feedback performance with few reward labels across a variety of domains and offline-RL backbones, yielding an effective method for feedback-efficient reinforcement learning.
\end{abstract}

\section{Introduction}
\label{sec:intro}

Modern large-scale sequential-decision problems pose a unique challenge: data can often be generated, logged, or collected at scale, but obtaining evaluative feedback for policy learning remains costly. In RLHF, models can generate many candidate responses, while feedback requires human judgment or a separate feedback oracle~\citep{christiano2017preferences,ouyang2022instructgpt,bai2022constitutional}; in drug discovery settings, generative models can propose large candidate sets, while validation requires expensive simulation or experimental evaluation~\citep{reymond2015chemical,gomez2018automatic,jin2018junction,popova2018deep,dimasi2016innovation}; and in offline RL, large reward-free transition datasets may be available while reward annotation remains limited~\citep{chaudhari2026data, yu2022unlabeled, hu2023pds}. Thus, in learning from collected data, the bottleneck often shifts from data collection to reward acquisition. When only a small fraction of the available data can be reward-labeled, the central question is \emph{which} subset to label so that learning from the resulting partially reward-labeled dataset yields a high-performing policy.

\looseness-1
Existing work on reinforcement learning under limited feedback (RLLF) studies reward selection in a setting where queried rewards are consumed by a policy learner rather than retained for reward-model training~\citep{chaudhari2026data}. This makes reward selection a question of which queried rewards directly improve policy learning. A complementary line of work on partially labeled offline RL shows that reward modeling is a useful way to leverage retained labels: rewards observed on a subset of transitions can be used to fit a reward model, which annotates the remaining offline dataset before policy optimization~\citep{yu2022unlabeled,hu2023pds,lai2025kernelpds}. However, these methods assume that the labeled subset is passively available. They do not address the question of which additional transitions should be selected for reward labeling. 
Information-directed reward learning (IDRL) is the closest related approach, as it actively queries feedback to learn a reward model for policy optimization. However, it acquires feedback online to distinguish the returns of candidate high-performing policies, rather than to determine which transitions in an offline dataset should be reward-labeled~\citep{lindner2021information}.
This leaves open the problem of \emph{retained-label reward selection}: actively selecting a small subset of transitions from a fixed, offline dataset for reward labeling, so that a reward model trained on those labels can annotate the remaining transitions and enable learning a high-performing policy from the resulting dataset.

\begin{figure}[t]
    \vspace{-2mm}
    \centering
    \includegraphics[width=\textwidth]{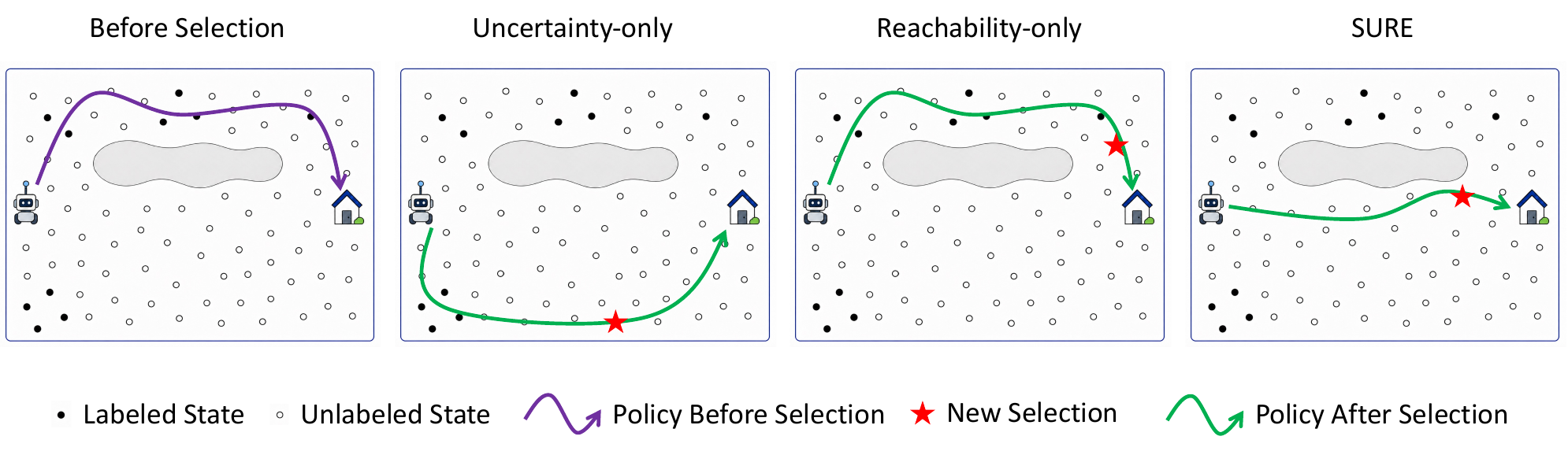}
    \caption{
    Illustrative retained-label reward-selection example in a navigation task. The agent must reach the goal while avoiding a central obstacle. With the initial reward labels, the policy learned from the current reward model takes a long detour, while a more direct path remains possible.
    Uncertainty-only selection chooses a state where the reward model is uncertain, but the selected state may be far from trajectories that lead to high return.
    Reachability-only selection chooses a state that can lead toward regions assigned high value by the current reward model, but this estimate can be misleading when the reward model is inaccurate.
    \SURE{} combines both signals, selecting for labeling a reward that is uncertain under the current reward model and lies on a reachable path toward currently high-value states, yielding a better policy with fewer reward labels.
    }
    \label{fig:illustration}
    \vspace{-8mm}
\end{figure}

\looseness-1
In this retained-label setting, selected rewards affect policy learning through the reward model they induce. A natural way to improve such a model is to label rewards that reduce uncertainty in the reward model, but reducing uncertainty at different states can have very different effects on the performance of the learned policy. We formalize this intuition by showing that the performance gap between the policy learned from a partially reward-labeled dataset and the true-reward optimal policy is upper bounded by two terms: one determined by the fixed offline transition dataset, and one given by reward uncertainty under the state distribution induced by the true-reward optimal policy. Since reward selection cannot alter the fixed transition data, it can only reduce the latter term. 
If the state occupancy of an optimal policy were known, the rule would be simple: select rewards that reduce uncertainty in states with high occupancy under that policy. Since this information is unavailable during selection, we introduce \emph{Successor-Guided Uncertainty REduction} (\SURE{}): the policy learned from the current reward model provides an evolving estimate of which states are valuable, while successor features are used to identify candidate states from which those states are reachable. \SURE{} selects rewards where reachability and uncertainty meet, i.e., states that remain uncertain and are reachable precursors to currently high-value states. \Cref{fig:illustration} illustrates this uncertainty--reachability tradeoff.

In this work, our contributions are:
\begin{enumerate}[leftmargin=*,topsep=0pt,itemsep=0pt]
    \item Formulate \emph{retained-label reward selection}, an offline RLLF setting where selected rewards are retained for reward-model learning and used to annotate the remaining offline data for policy learning. This extends RLLF to retained reward labels and continuous state-action domains.

    \item Establish a bound-guided acquisition principle for retained-label selection. By separating the performance gap into a fixed offline-data term and a label-dependent reward-uncertainty term, our analysis identifies the ideal selection target: reducing uncertainty on states visited by the true-reward optimal policy.

    \item Introduce \SURE{}, an adaptive reward-selection rule that implements this principle by combining reward-model uncertainty with successor-guided reachability to currently high-value states.

    \item Analyze and evaluate the reward-label efficiency of \SURE{} through theory and experiments. Our analyses relate selection quality to coverage mismatch and effective dimension. Experiments across a variety of domains and offline-RL backbones show that \SURE{} reaches strong performance with fewer reward labels than competing selection strategies.
\end{enumerate}

\section{Related Work}
\label{sec:related}

\looseness-1
\paragraph{Reward selection and active reward acquisition.}
When only a limited amount of reward feedback can be obtained, different problem settings involve learners that can query different objects. Active reward learning and preference-based RL query comparisons, demonstrations, rankings, or other types of human feedback to identify a latent reward function~\citep{daniel2015active,sadigh2017active,biyik2020asking,wilde2020active}. 
Information-directed reward learning (IDRL) is closer to our motivation: these methods actively acquire feedback to learn a reward model for policy optimization~\citep{lindner2021information}. However, IDRL acquires feedback online, selecting queries expected to rapidly identify which candidate high-performing policy achieves the highest return, whereas our setting focuses on selecting which transitions in a fixed offline dataset should receive reward labels. 
Methods for active RL with costly rewards and partially observable reward models also query reward observations while interacting with the environment, intertwining reward acquisition with exploration and data collection~\citep{krueger2020active,schulze2018active,tucker2023bandits,parisi2024monitored,parisi2024beyond}. 
These directions all study how to spend limited feedback, but their queried objects and learning protocols differ from retained-label offline reward selection.
Closest to our setting is RLLF~\citep{chaudhari2025which,chaudhari2026data}, which formulates reward selection from a reward-free offline dataset under a limited labeling budget. We build on this formulation but retain queried rewards for reward-model learning. The resulting reward-uncertainty signals guide label selection directly, avoiding the additional policy-evaluation cost of the evaluator used in RLLF.

\looseness-1
\paragraph{Partially labeled offline RL.}
Partially labeled offline RL studies policy learning from offline datasets in which only a subset of transitions is annotated with rewards, while the remaining transitions are reward-free. The UDS method~\citep{yu2022unlabeled} studies this setting directly and shows that simply assigning zero rewards to unlabeled transitions can be surprisingly effective, despite the bias introduced by incorrect reward labels. Another strategy is to infer missing reward signals from the labeled subset and use the learned reward model to annotate reward-free transitions before applying standard offline RL \citep{konyushkova2020semi,zolna2020offline,romeo2024imputed}. 
The PDS method~\citep{hu2023pds} takes a pessimistic approach to offline RL with partially labeled datasets, learning a conservative reward estimate from the labeled subset and using the full transition dataset for policy learning.
Kernelized variants extend this partially labeled offline-RL framework beyond the original linear setting \citep{lai2025kernelpds}. Related offline policy-learning settings consider additional forms of incomplete supervision, such as observation-only data with partially labeled rewards, missing actions, low-quality demonstrations, or limited coverage \citep{li2023mahalo,nishimori2025domain}.

\section{Problem Formulation and Preliminaries}
\label{sec:prelim}

\paragraph{Preliminaries.}
An MDP is a tuple $\mathcal M=(\mathcal S,\mathcal A,p,r,\gamma,\eta)$, where $\mathcal S$ is the state space, $S_t$ is the state at time $t \in \{0,1,\dotsc\}$, $\mathcal A$ is the action space, $A_t$ is the action at time $t$, $p(s'\mid s,a):=\Pr(S_{t+1}=s'\mid S_t=s,A_t=a)$ is the transition function, $r(s,a)\in[0,r_{\max}]$ is the reward function, $\gamma\in[0,1)$ is the discount factor, and $\eta(s):=\Pr(S_0=s)$ is the initial-state distribution. Specifically, we consider linear MDPs for our theoretical exposition, defined as follows.

\begin{definition}[Linear MDP~\citep{yang2019reinforcement,jin2020provably}]
\label{def:linear_mdp}
An MDP $\mathcal M=(\mathcal S,\mathcal A,p,r,\gamma,\eta)$ is a linear MDP if there exists a feature map $\phi:\mathcal S\times\mathcal A\to\mathbb R^d$ with $\|\phi(s,a)\|_2\le 1$ for all $(s,a)$, an unknown vector-valued measure $\omega=(\omega_1,\ldots,\omega_d)$ over $\mathcal S$, and an unknown reward parameter $\theta^\star\in\mathbb R^d$ such that $p(\cdot\mid s,a)=\phi(s,a)^\top\omega(\cdot)$ and $r(s,a)=\phi(s,a)^\top\theta^\star$ for all $(s,a)$. We assume $\|\theta^\star\|_2\le S_r$, where $S_r>0$ bounds the reward-parameter norm.
\end{definition}

Given a policy $\pi(a\mid s):=\Pr(A_t=a\mid S_t=s)$, $J_r(\pi):=\mathbb E_\pi[\sum_{t\ge 0}\gamma^t r(s_t,a_t)]$ denotes its discounted expected return under reward function $r$. Consider this MDP, but now with a different reward function $r'$; we denote $\pi$'s discounted expected return under $r'$ by $J_{r'}(\pi)$. Let $\pi_r^\star\in\arg\max_\pi J_r(\pi)$ be an optimal policy under reward $r$. The discounted state--action occupancy measure of $\pi$ is $d^\pi(s,a):=(1-\gamma)\sum_{t\ge 0}\gamma^t\Pr((s_t,a_t)=(s,a);\pi)$, which satisfies the relation $J_r(\pi)=(1-\gamma)^{-1}\mathbb E_{(S,A)\sim d^\pi}[r(S,A)]$~\citep{sutton2018reinforcement}. A closely related concept is \textit{successor features}, which encode the expected discounted occupancy of future state-action features.

\begin{definition}[Successor Features~\citep{dayan1993successor,barreto2017successor}]
\label{def:successor_feature}
For a policy $\pi$, the successor features of a state--action pair $(s,a)$ is defined as:
\begin{equation}
\psi^\pi(s,a)
:=
\mathbb E_\pi\!\left[
\sum_{t\ge 0}\gamma^t\phi(S_t,A_t)
\,\middle|\,
(S_0,A_0)=(s,a)
\right].
\label{eq:successor_feature}
\end{equation}
\end{definition}

Successor features represent the discounted future occupancy of the feature $\phi$ under a given policy. They summarize which feature directions trajectories from $(s,a)$ are expected to visit.

\subsection{Retained-Label Reward Selection}
\label{subsec:retained_label_selection}

We adopt the reward-selection formulation from~\citet{chaudhari2026data} (RLLF). An offline dataset $\mathcal D=\{(s_i,a_i,s_i')\}_{i=1}^N$ is obtained by the interaction of a data-collecting policy $\pi_D$ with $\mathcal M$, and contains no rewards. Let $\mathcal S_{[B]}$ denote the states in $\mathcal D$ that are reward-labeled: for samples in $\mathcal D$ where $S_t \in \mathcal S_{[B]}$, reward labels are assigned; the remaining samples in $\mathcal D$ are unlabeled. Given a labeling budget $B$, a reward-selection strategy is a map $\mathcal Q^{(B)}:\mathcal D\to\{T\subseteq\mathcal S:|T|=B\}$. The states selected for reward labeling are $\mathcal S_{[B]}=\mathcal Q^{(B)}(\mathcal D)$. 
Unlike RLLF's original setting, where queried reward labels for states in $\mathcal S_{[B]}$ are not retained, in our setting each transition $(s_i,a_i,s_i')\in\mathcal D$, where $s_i\in\mathcal S_{[B]}$, receives a reward label $r_i=r(s_i,a_i)$, and the resulting labeled set, $\mathcal Y_{[B]}:=\{(s_i,a_i,r_i,s_i'):(s_i,a_i,s_i')\in\mathcal D,\ s_i\in\mathcal S_{[B]}\}$, is retained for reward-model learning. 
In our setting, the selected rewards are retained and used to learn a policy $\pi^{[B]}$ through the reward-modeling and offline-RL pipeline described next. Our reward-selection objective, in particular, is to identify a querying strategy $\mathcal Q^{(B)\star} \in \arg\max_{\mathcal Q^{(B)}} P(\mathcal Q^{(B)})$, where $P(\mathcal Q^{(B)}):=J_r(\pi^{[B]})$ is the return of the policy learned from the retained labels selected by $\mathcal Q^{(B)}$.

\subsection{Reward Modeling for Partially Labeled Offline RL}
\label{subsec:plorl_backbone}

Given a retained labeled set $\mathcal Y$, we follow the Provable Data Sharing (PDS) framework~\citep{hu2023pds}: fit a conservative reward model from $\mathcal Y$, use it to assign rewards to the remaining unlabeled transitions in $\mathcal D$, and run a pessimistic offline-RL on the resulting annotated dataset. Under the linear reward model, we fit the best linear reward parameter from the labeled transitions using $\ell_2$ regularization, commonly referred to as ridge regression: 
$
\hat\theta_{\mathcal Y}
\in
\arg\min_{\theta\in\mathbb R^d}
\sum_{(s_i,a_i,r_i,s_i')\in\mathcal Y}
\bigl(\phi(s_i,a_i)^\top\theta-r_i\bigr)^2
+
\lambda\|\theta\|_2^2 .
\label{eq:reward_ridge_objective}
$

Let 
$\Lambda_{\mathcal Y}:=\lambda I+\sum_{(s_i,a_i,r_i,s_i')\in\mathcal Y}\phi(s_i,a_i)\phi(s_i,a_i)^\top$
denote the labeled reward-design matrix. The ridge solution is
$\hat\theta_{\mathcal Y}=\Lambda_{\mathcal Y}^{-1}\sum_{(s_i,a_i,r_i,s_i')\in\mathcal Y}\phi(s_i,a_i) r_i$.
Using this estimate, PDS constructs a pessimistic conservative reward model, 
\[
\tilde r_{\mathcal Y}(s,a)
:=
\bigl[\phi(s,a)^\top\hat\theta_{\mathcal Y}
-
\alpha_{\mathcal Y}\sigma_{\mathcal Y}(s,a)
\bigr]_+,
\]
where $[x]_+:=\max\{x,0\}$, and
\[
\sigma_{\mathcal Y}(s,a)
:=
\sqrt{\phi(s,a)^\top\Lambda_{\mathcal Y}^{-1}\phi(s,a)},\qquad
\alpha_{\mathcal Y}
:=
\sqrt{\lambda}\,S_r+
\sigma_r\sqrt{2\log(1/\delta)+d\log(1+|\mathcal Y|/\lambda)} .
\]
Here, $\sigma_{\mathcal Y}(s,a)$ is the reward-model uncertainty induced by the retained labeled set $\mathcal Y$, and $\alpha_{\mathcal Y}$ is the corresponding confidence width, which depends on the budget $B$ and the confidence tolerance $\delta\in(0,1)$. We take $\lambda\ge 1$, so $\Lambda_{\mathcal Y}\succeq I$ and $\sigma_{\mathcal Y}(s,a)\le\|\phi(s,a)\|_2\le 1$. Appendix~\ref{app:pds_pipeline} provides the full derivation.

By the self-normalized concentration inequality of \citet{abbasi2011linear} and the nonnegative-reward clipping in Definition~\ref{def:linear_mdp}, with probability at least $1-\delta$, the following holds simultaneously for all labeled sets $\mathcal Y$ encountered by the learner and for all $(s,a)\in\mathcal S\times\mathcal A$: 
\begin{equation}
0
\le
r(s,a)-\tilde r_{\mathcal Y}(s,a)
\le
2\alpha_{\mathcal Y}\sigma_{\mathcal Y}(s,a).
\label{eq:pessimistic_reward_gap}
\end{equation}
The derivation is provided in Appendix~\ref{app:reward_confidence}.
Given the reward estimate $\tilde r_{\mathcal Y}$, the downstream learner runs a pessimistic offline-RL algorithm on $\mathcal D$. 
We use the PEVI-style backbone~\citep{jin2021pessimism} as a black box providing the following high-probability guarantee: for any pessimistic reward estimate $\tilde r\le r$,
\begin{equation}
J_{\tilde r}(\pi^\star_{\tilde r})
-
J_{\tilde r}(\hat\pi_{\tilde r})
\le
\Gamma_{\rm off}(\mathcal D,\delta),
\label{eq:offline_backbone_gap}
\end{equation}
\looseness-1
where
$
\Gamma_{\rm off}(\mathcal D,\delta)
:=
\frac{2\beta_{\rm off}}{1-\gamma}\,
\max_{(s,a)\in\mathcal D}
\sqrt{\phi(s,a)^\top \Lambda(\mathcal D)^{-1}\phi(s,a)},
$
and $\beta_{\rm off}=\widetilde O(d/(1-\gamma))$ is the offline confidence width. The full derivation is given in Appendix~\ref{app:gamma_off_form}. Crucially, this upper bound depends only on the fixed transition dataset $\mathcal D$ and confidence tolerance $\delta$, not on which transitions are reward-labeled.
This motivates the reward-selection question studied next: which rewards should be queried/labeled (and retained) so that the resulting reward model improves policy learning the most?

\section{Successor-Guided Uncertainty Reduction}
\label{sec:method}

\looseness-1
We introduce \emph{Successor-Guided Uncertainty REduction} (\SURE{}), an adaptive reward-selection method for the retained-label RLLF setting. Following RLLF's iterative selection protocol, \SURE{} updates the retained labeled dataset after each queried reward, refits a pessimistic estimate of the reward model, and uses the resulting policy and value estimates to choose the next label. In round $b\in\{0,\ldots,B-1\}$, \SURE{} fits the current pessimistic reward model $\tilde r_b$ from $\mathcal Y^{[b]}$, runs the offline-RL learner to obtain $\pi^{[b]}$, selects an unlabeled transition $(s_i,a_i,s_i')\in\mathcal D$, queries $r_i=r(s_i,a_i)$, and updates $\mathcal Y^{[b+1]}=\mathcal Y^{[b]}\cup\{(s_i,a_i,r_i,s_i')\}$. The score guiding this selection is derived below from a \textit{label-sensitive bound} (Thm.~\ref{thm:label_sensitive}) and then approximated through successor-guided reachability and reward-model uncertainty metrics (Section~\ref{subsec:successor_guidance}).

\subsection{From the Suboptimality Bound to an Acquisition Objective}
\label{subsec:bound_to_objective}

We first derive a \textit{label-sensitive upper bound on policy suboptimality}. The bound separates the contribution of the fixed offline transition dataset from the contribution of the rewards selected so far.

\begin{theorem}[Label-sensitive suboptimality decomposition]
\label{thm:label_sensitive}
Assume that the MDP is linear, as in Definition~\ref{def:linear_mdp}, that the retained labels induce a conservative reward estimate satisfying the reward-confidence bound (Eq.~\eqref{eq:pessimistic_reward_gap}), and that the pessimistic offline-RL algorithm satisfies the stated offline-learning guarantee (Eq.~\eqref{eq:offline_backbone_gap}). 
For any strategy with budget $B$, let $\mathcal Q^{(b)}$ denote its first $b$ selections, let $\mathcal Y^{[b]}$ denote the retained labeled set resulting from those selections, and write $\sigma_b:=\sigma_{\mathcal Y^{[b]}}$ and $\alpha_b:=\alpha_{\mathcal Y^{[b]}}$. Then, with probability at least $1-2\delta$, for every round $b\in\{0,\ldots,B\}$,
\begin{equation}
\label{eq:method_decomp}
\SubOpt(\mathcal Q^{(b)})
\le
\underbrace{\Gamma_{\mathrm{off}}(\mathcal D,\delta)}_{\textcolor{blue}{\text{fixed offline-data term}}}
+
\underbrace{
\frac{2\alpha_b}{1-\gamma}
\sqrt{
\mathbb{E}_{(s,a)\sim d^{\pi_r^\star}}\!\left[\sigma_b(s,a)^2\right]
}
}_{\textcolor{red!70!black}{\text{label-dependent reward-uncertainty term}}},
\end{equation}
where $\SubOpt(\mathcal Q^{(b)}) := J_r(\pi_r^\star)-J_r(\pi^{[b]})$ is the suboptimality of $\pi^{[b]}$, the policy learned from the retained labeled set $\mathcal Y^{[b]}$ induced by $b$ selections.
\end{theorem}

\looseness-1
The proof is in Appendix~\ref{app:proof_label_sensitive}. Thm.~\ref{thm:label_sensitive} shows that the suboptimality bound for the policy induced by a querying strategy $\mathcal Q^{(b)}$ can be tightened by minimizing a reward-uncertainty term that depends on the optimal occupancy $d^{\pi_r^\star}$ under the true reward function. Because $d^{\pi_r^\star}$ is unknown during selection, we need a computable surrogate (see Section~\Cref{subsec:successor_guidance}). Here, we first derive the corresponding selection objective for an arbitrary reference measure $\mu$.

\looseness-1
Given a retained labeled set $\mathcal Y^{[b]}$ and reference measure $\mu$, define the \textit{weighted uncertainty objective} as
\begin{equation}
\label{eq:def_phi}
\Phi(\mathcal Y^{[b]};\mu)
:=
\mathbb E_{(s,a)\sim\mu}[\sigma_b(s,a)^2].
\end{equation}
This objective measures the squared reward uncertainty averaged over $\mu$. Thm.~\ref{thm:label_sensitive} can then be written as
\begin{equation}
\label{eq:method_phi_bound}
\SubOpt(\mathcal Q^{(b)})
\le
\Gamma_{\mathrm{off}}(\mathcal D,\delta)
+
\frac{2\alpha_b}{1-\gamma}
\sqrt{\Phi(\mathcal Y^{[b]};d^{\pi_r^\star})}.
\end{equation}

\looseness-1
Since only the second term in Eq.~\eqref{eq:method_phi_bound} depends on the selected rewards, we can use $\Phi(\mathcal Y^{[b]};\mu)$ as the objective guiding reward selection. For a fixed reference measure $\mu$, we define the \textbf{one-step acquisition function} as the uncertainty reduction in this objective obtained by reward-labeling a candidate $(s,a)$:
\begin{equation}
\label{eq:def_delta}
\Delta(s,a;\mathcal Y^{[b]},\mu)
:=
\Phi(\mathcal Y^{[b]};\mu)
-
\Phi(\mathcal Y^{[b]}\cup\{(s,a)\};\mu).
\end{equation}
\looseness-1
We now show (Prop.~\ref{prop:exact_gain}) that this marginal gain admits a closed form under the linear-MDP assumption.

\begin{proposition}[Exact one-step marginal gain]
\label{prop:exact_gain}
Let $\Sigma(\mu):=\mathbb E_{(s',a')\sim\mu}[\phi(s',a')\phi(s',a')^\top]$. For any reference measure $\mu$ and candidate $(s,a)$,
\begin{equation}
\label{eq:exact_gain}
\Delta(s,a;\mathcal Y^{[b]},\mu)
=
\frac{
\bigl\|\Sigma(\mu)^{1/2}\Lambda_b^{-1}\phi(s,a)\bigr\|_2^2
}{
1+\sigma_b(s,a)^2
}.
\end{equation}
\end{proposition}

The proof is given in Appendix~\ref{app:proof_exact_gain}. Equation~\eqref{eq:exact_gain} shows that the one-step acquisition function $\Delta$ accounts for both the reference measure and the current labeled set: it favors candidates that reduce uncertainty in directions emphasized by $\mu$, while downweighting feature directions already well represented by the current labels. For fixed $\mu$, maximizing $\Delta(\cdot;\mathcal Y^{[b]},\mu)$ gives the largest one-step decrease of the bound in~\eqref{eq:method_phi_bound}.
Appendix~\ref{app:delta_monotonicity} formalizes this monotonicity. We next construct a computable reference measure to replace the unavailable optimal occupancy $d^{\pi_r^\star}$.

\subsection{Successor-Guided Surrogate Occupancy}
\label{subsec:successor_guidance}

At iteration $b$, a pessimistic reward model $r_b$ is learned from the retained labels $\mathcal Y^{[b]}$. 
The policy $\pi^{[b]}$ learned from the dataset annotated by $r_b$ provides value estimates $Q_b$ on the offline dataset. We use these estimates to construct a computable proxy for the optimal policy's occupancy: following RLLF's intuition that useful rewards should help the agent remain on, or recover toward, high-return behavior, we treat the highest-valued transitions as temporary targets, or subgoals, and use successor features to score which unlabeled candidates can reach them over multiple steps.

Let $T_b\subset\mathcal D$ be the top-$K$ transitions in the offline dataset ranked by $Q_b(s,a)$. 
For a candidate $(s_i,a_i)$ and target $(s',a')$, define the successor overlap $h_b((s_i,a_i),(s',a')):=\psi^{\pi^{[b]}}(s_i,a_i)^\top\phi(s',a')$. This overlap is large when trajectories starting from $(s_i,a_i)$ visit feature directions similar to the target under $\pi^{[b]}$. Averaging over the target set gives
\begin{equation}
\label{eq:reachability_score}
H_b(s_i,a_i)
:=
\frac{1}{|T_b|}
\sum_{(s',a')\in T_b}
h_b((s_i,a_i),(s',a')) .
\end{equation}
We normalize the positive scores over unlabeled candidates to obtain a successor-guided distribution $\bar H_b$; the explicit normalization is given in Appendix~\ref{app:normalized_reachability}. To avoid over-concentration when the current value estimates are inaccurate, we mix this distribution with the empirical behavior distribution $\hat d_{\mathrm{beh}}^{(b)}$ over unlabeled samples:
\begin{equation}
\label{eq:mixed_measure}
\rho_b
=
\beta_b \hat d_{\mathrm{beh}}^{(b)}
+
(1-\beta_b)\bar H_b,
\qquad
\beta_b\in[0,1].
\end{equation}
We refer to $\rho_b$ as the successor-guided reachability measure.

The measure $\rho_b$ is the computable reference measure used in place of the unavailable optimal occupancy $d^{\pi_r^\star}$. The behavior mixture acts as a coverage safeguard: it prevents successor guidance from concentrating entirely on the current value estimate and preserves feature directions covered by the empirical behavior distribution.
Appendix~\ref{app:theoretical_analysis} provides the corresponding theoretical analysis: reference measures constructed from the offline dataset transfer to the original optimal-occupancy objective, with guarantees controlled by coverage of the relevant feature directions.

In principle, one could plug the reachability measure $\rho_b$ into the exact gain and select
$\arg\max_{(s_i,a_i)\in\mathcal D\setminus\mathcal S_b}
\Delta(s_i,a_i;\mathcal Y^{[b]},\rho_b)$.
However, evaluating this exact gain for every candidate requires applying
$\Sigma(\rho_b)^{1/2}\Lambda_b^{-1}$ to each candidate feature $\phi(s_i,a_i)$, so that the score accounts for uncertainty reduction in feature directions emphasized by $\rho_b$. Forming this matrix and scoring all unlabeled candidates has per-round computational complexity
$O(d^3 + (|\mathcal D|-b)d^2)$ (Appendix~\ref{app:complexity}). To avoid this candidate-wise matrix-vector scoring, we define the computable \SURE{} score as
\begin{equation}
\label{eq:swvr}
\SUREscore_b(s_i,a_i)
=
\rho_b(s_i,a_i)
\frac{\sigma_b(s_i,a_i)^2}{1+\sigma_b(s_i,a_i)^2}.
\end{equation}

The first factor is the successor-guided reachability measure, which prioritizes candidates that can reach currently high-value targets while maintaining behavior-policy coverage. The second factor is the local uncertainty-reduction term, which prioritizes candidates that remain uncertain under the current reward model, with the same normalization used in Proposition~\ref{prop:exact_gain}. \SURE{} combines these factors to select rewards that are both reachable to currently high-value targets and informative for reward-model learning, thereby targeting uncertainty reductions that are more likely to improve the learned policy. Section~\ref{sec:qualitative_ablation} shows that either factor alone can fail: \textsc{Reach}, which selects only by the reachability factor, can fail when current value estimates are inaccurate, while \textsc{Uncertainty}, which selects only by the uncertainty factor, can fail when high uncertainty lies in regions unrelated to high-return behavior. Appendix~\ref{app:algorithm} gives the full adaptive procedure.

\vspace{-2mm}
\section{Experiments}
\label{sec:experiments}

We evaluate SURE in the retained-label RLLF setting considered in Section~\ref{sec:prelim}, where the learner is given a fixed reward-free offline transition dataset $\mathcal D$ and can acquire rewards for only $B$ selected transitions. Our experiments address five questions: (\textbf{Q1} Section~\ref{sec:linear_mdp}) when the linear-MDP assumptions of Section~\ref{sec:prelim} hold by construction, does the computable SURE score retain the label efficiency of the exact one-step-gain rule it approximates; (\textbf{Q2} Section~\ref{sec:main_comparison}) does SURE remain label-efficient beyond this setting, in continuous-control domains where the feature map is learned and the linear-MDP structure need not hold, and how does it compare with other reward-selection strategies across budgets; (\textbf{Q3} Section~\ref{sec:feedback_percentage}) how much feedback each strategy requires to approach the full-feedback performance obtained when rewards are available for all transitions in $\mathcal D$; (\textbf{Q4} Section~\ref{sec:qualitative_ablation}) how each component of SURE contributes to acquisition; and (\textbf{Q5} Section~\ref{sec:backbone}) whether the gains of SURE are specific to the IQL backbone or persist when the downstream offline-RL learner is replaced.

We evaluate on five domains with different acquisition challenges: LinearMDP, a finite linear MDP in which the assumptions of Section~\ref{sec:prelim} hold by construction (Section~\ref{sec:linear_mdp}); TwoRooms, Maze2d-Medium, and Maze2d-Large, three navigation tasks with bottleneck and goal-reaching rewards; and HalfCheetah, a high-dimensional locomotion task~\citep{brockman2016gym}. For each domain, all strategies use the same fixed reward-free dataset $\mathcal D$: we collect the TwoRooms dataset ourselves and use the D4RL datasets~\citep{fu2020d4rl} for Maze2d-Medium, Maze2d-Large, and HalfCheetah.

Our empirical comparison focuses on reward-selection strategies: SURE selects rewards using the acquisition score in equation~\eqref{eq:swvr}, and we compare it against alternative rules for selecting the rewards. \textsc{Uniform} samples labels uniformly from the unlabeled pool. \textsc{RLLF-Guided} adapts the guided heuristic of RLLF~\citep{chaudhari2026data}, which biases selection toward one-step predecessors of states estimated to have high value under the current learner. \textsc{IDRL}~\citep{lindner2021information} adapts the information-directed acquisition criterion to the fixed offline dataset by estimating policy-return information with FQE. \textsc{Reach} uses only the successor-guided relevance factor in equation~\eqref{eq:swvr}, while \textsc{Uncertainty} uses only the uncertainty-reduction factor.
After rewards are selected, every method follows the retained-label PDS pipeline described in Section~\ref{sec:prelim}: the acquired labels are used to fit a conservative reward model, which assigns rewards to the unlabeled portion of $\mathcal D$; IQL~\citep{kostrikov2021iql} is then run on the resulting reward-labeled dataset to produce a policy (Section~\ref{sec:backbone} replaces IQL with other offline-RL backbones).

We report the return of learned policies as a function of the absolute feedback budget $B$ in Section~\ref{sec:main_comparison} and the percentage feedback $B/|\mathcal S_{\mathcal D}|$ in Section~\ref{sec:feedback_percentage}, where $\mathcal S_{\mathcal D}:=\{s_i:(s_i,a_i,s_i')\in\mathcal D\}$ denotes the set of distinct states in $\mathcal D$. We also include a \emph{full-feedback reference}, corresponding to percentage feedback equal to one, i.e., $B=|\mathcal S_{\mathcal D}|$. Results are averaged over 50 trials for each strategy. Full domain and dataset details, baseline descriptions, implementation details, hyperparameters, and evaluation protocols are provided in Appendix~\ref{app:experiments}.

\subsection{Validation under Exact Linear MDPs}
\label{sec:linear_mdp}

We first evaluate SURE in a setting where the theoretical assumptions hold exactly. We construct finite linear MDPs using the simplex-feature construction of \citet{jin2020provably}: each state--action pair is assigned a known $8$-dimensional feature vector on the probability simplex, its transition distribution is the corresponding convex combination of eight base next-state distributions, and its reward is linear in the same feature vector. Both the reward and the transition kernel are therefore linear in a fixed feature map $\phi$, which is provided to every method. We consider environments with $32$ and $48$ states, four actions, and low-exploration offline pools of $4{,}096$ transitions. Because $\phi$ is known and the problem is small, the exact one-step gain of Proposition~\ref{prop:exact_gain} can be computed for every candidate, so we compare SURE with \textsc{Exact Gain}, which selects the candidate maximizing equation~\eqref{eq:exact_gain}, and with \textsc{Uniform}. All methods share the same four uniformly selected warm-start labels, counted toward the budget, and results are averaged over $30$ paired seeds.

\begin{table}[b]
\centering
\caption{Number of reward labels required to reach $80\%$ / $90\%$ / $95\%$ of the matched full-feedback return in exact linear MDPs, averaged over $30$ paired seeds. Full feedback corresponds to labeling all $4{,}096$ transitions in the offline pool.}
\label{tab:linear_mdp}
\begin{tabular}{lccc}
\toprule
Environment & \textsc{SURE} & \textsc{Exact Gain} & \textsc{Uniform} \\
\midrule
$|\mathcal S|=32$ & \textbf{8 / 12 / 12} & \textbf{8 / 12 / 12} & 20 / 20 / 36 \\
$|\mathcal S|=48$ & \textbf{8 / 12 / 12} & \textbf{8 / 12 / 12} & 12 / 68 / 132 \\
\bottomrule
\end{tabular}
\vspace{-3mm}
\end{table}

Table~\ref{tab:linear_mdp} shows that SURE and \textsc{Exact Gain} require identical numbers of labels at every threshold in both environments. With $48$ states, both reach $90\%$ and $95\%$ of the full-feedback return with $12$ labels, whereas \textsc{Uniform} requires $68$ and $132$ labels, respectively. Thus, when the theoretical assumptions hold, replacing the exact one-step gain with the computable SURE score does not sacrifice label efficiency.

\subsection{Label Efficiency Across Domains}
\label{sec:main_comparison}

\begin{figure}[t]
    \centering
    \includegraphics[width=\textwidth]{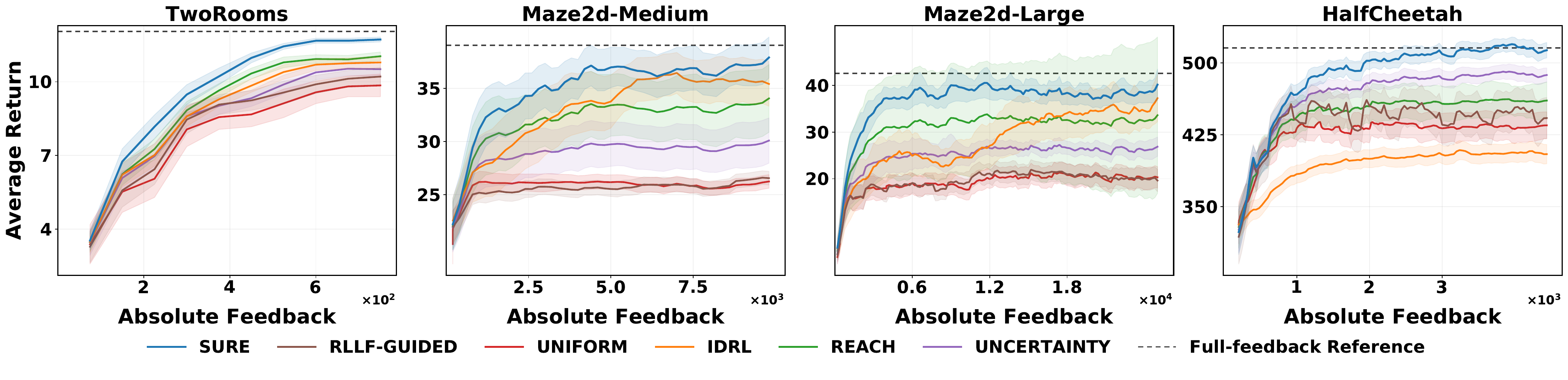}
    \vspace{-3mm}
    \caption{Performance comparison across domains under limited feedback. Return is shown as a function of the absolute feedback budget $B$. We compare \textsc{SURE} with \textsc{Uniform}, \textsc{Reach}, \textsc{Uncertainty}, \textsc{IDRL}, and \textsc{RLLF-Guided}. Solid lines show the mean over 50 trials and shaded regions show the standard error. The dashed line indicates the \emph{full-feedback reference}.}
    \label{fig:main_comparison}
    \vspace{-4mm}
\end{figure}

In continuous-control tasks, however, a feature map under which rewards and transitions are exactly linear is generally unknown and must be learned from data. We therefore test whether the conclusions of Section~\ref{sec:linear_mdp} carry over to four continuous-control domains in which $\phi$ is learned rather than given. Figure~\ref{fig:main_comparison} compares SURE with the reward-selection baselines described above. The solid curve shows the mean return over 50 trials, and the shaded region shows the standard error. The dashed line indicates the full-feedback offline-RL reference, obtained by training IQL when rewards are available for all transitions in $\mathcal D$.

The results show that the relative performance of baselines varies across domains, while SURE remains consistently strong across all settings. Below, we highlight the main empirical findings:
\begin{itemize}
    \item \textbf{SURE is the strongest overall strategy.} Across all four domains, SURE achieves the best or near-best returns over the feedback range. The strongest competing baseline changes across domains, but SURE is the only method that remains consistently strong across the full evaluation suite.

    \item \textbf{Single-factor acquisition is not sufficient.} The better single-factor baseline depends on the domain. In navigation tasks, \textsc{Reach} is often competitive because the most useful labels are those that help identify trajectories leading toward the goal. In HalfCheetah, this spatial guidance is less informative: the higher-dimensional replay dataset makes reward uncertainty the main bottleneck, so \textsc{Uncertainty} becomes the stronger single-factor baseline. This domain dependence shows that neither reachability nor uncertainty alone is a reliable acquisition rule.

    \item \textbf{Information-directed querying depends on reliable return estimates.} \textsc{IDRL} is competitive in the maze domains, where low-dimensional geometry and goal-reaching rewards make policy-return differences easier to estimate from limited rewards. Its performance degrades in HalfCheetah, where the higher-dimensional dataset makes the information-directed acquisition signal less reliable.

    \item \textbf{Multi-step reachability improves over one-step guidance.} \textsc{RLLF-Guided} often remains close to \textsc{Uniform}, suggesting that one-step predecessor guidance alone is too local for these offline datasets. In contrast, \textsc{Reach} is stronger in navigation domains, indicating that successor-guided multi-step reachability provides a more useful acquisition signal than selecting only immediate predecessors of currently high-value states.

    \item \textbf{SURE combines the useful signals without choosing one in advance.} The multiplicative acquisition score requires a candidate to be both relevant to promising behavior and informative for reward-model learning. As a result, SURE emphasizes reachability when useful trajectories are easier to identify from the geometry of the task, and emphasizes uncertainty when the relevant behaviors are harder to localize.
\end{itemize}

\vspace{-3mm}
\subsection{Feedback Needed to Reach Full-Feedback Performance}
\label{sec:feedback_percentage}

\begin{table}[t]
\centering
\caption{Percentage feedback needed to reach different fractions of the \emph{full-feedback reference}. The ``Fraction'' column specifies the target return as a fraction of the \emph{full-feedback reference}; ``Percentage feedback required'' reports the fraction of distinct states that must be labeled to reach that target. \textbf{Bold} indicates the smallest percentage feedback in each row.}
\label{tab:budget-to-threshold}
\setlength{\tabcolsep}{3.5pt}
\begin{tabularx}{\linewidth}{l>{\centering\arraybackslash}Xrrrrrr}
\toprule
\multirow{2}{*}[-1.8ex]{Domain}
& \multirow{2}{*}[-1.8ex]{Fraction}
& \multicolumn{6}{c}{Percentage feedback required} \\
\cmidrule(lr){3-8}
&
& \textsc{SURE}
& \makecell{\textsc{RLLF}\\\textsc{-Guided}}
& \textsc{Uniform}
& \textsc{IDRL}
& \textsc{Reach}
& \textsc{Uncertainty} \\
\midrule
\multirow{3}{*}{maze2d-medium}
  & 95\% & \textbf{0.20\%} & 1.74\% & 2.25\% & 0.32\% & 1.46\% & 1.96\% \\
  & 90\% & \textbf{0.12\%} & 1.45\% & 1.79\% & 0.26\% & 0.50\% & 1.47\% \\
  & 80\% & \textbf{0.05\%} & 0.94\% & 1.22\% & 0.11\% & 0.06\% & 0.50\% \\
\midrule
\multirow{3}{*}{maze2d-large}
  & 95\% & \textbf{0.15\%} & 3.28\% & 2.88\% & 0.64\% & 2.51\% & 2.79\% \\
  & 90\% & \textbf{0.09\%} & 2.92\% & 2.70\% & 0.62\% & 2.10\% & 2.52\% \\
  & 80\% & \textbf{0.06\%} & 2.42\% & 2.38\% & 0.40\% & 0.16\% & 2.12\% \\
\midrule
\multirow{3}{*}{halfcheetah}
  & 95\% & \textbf{0.71\%} & 6.64\% & 7.06\% & 6.18\% & 4.10\% & 2.49\% \\
  & 90\% & \textbf{0.45\%} & 2.55\% & 2.65\% & 2.22\% & 2.49\% & 0.59\% \\
  & 80\% & \textbf{0.30\%} & 0.31\% & 0.36\% & 2.20\% & 0.32\% & 0.31\% \\
\bottomrule
\end{tabularx}
\vspace{-6mm}
\end{table}

Table~\ref{tab:budget-to-threshold} converts the curves in Figure~\ref{fig:main_comparison} into the feedback required to reach fixed fractions of the \emph{full-feedback reference}. This view makes the cross-domain comparison more direct by normalizing for the size of each domain's state set: TwoRooms has $15$K transitions, Maze2d-Medium has $2$M, Maze2d-Large has $4$M, and HalfCheetah has $200$K.

The main takeaway is that SURE reaches the same performance levels with substantially less feedback. At the 95\% level, SURE requires the least feedback in every domain: $3.17\%$ in TwoRooms, $0.20\%$ in Maze2d-Medium, $0.15\%$ in Maze2d-Large, and $0.71\%$ in HalfCheetah. Compared with the strongest non-SURE method in each domain, this corresponds to roughly $1.6\times$--$4.3\times$ less feedback. The gap is even larger against mismatched single-factor baselines: \textsc{Uncertainty} requires about $18.6\times$ more feedback than SURE on Maze2d-Large, while \textsc{Reach} requires about $5.8\times$ more feedback than SURE on HalfCheetah.

Since SURE reaches the 95\% level with substantially less feedback than the baselines, Figure~\ref{fig:main_comparison} focuses on the feedback range in which SURE has largely converged. Some baselines continue to improve with larger budgets, but Table~\ref{tab:budget-to-threshold} shows that they require much more feedback to reach the same performance levels.

\subsection{Ablation Study of Reachability and Uncertainty}
\label{sec:qualitative_ablation}

\begin{figure}[t]
    \centering
    \includegraphics[width=\textwidth]{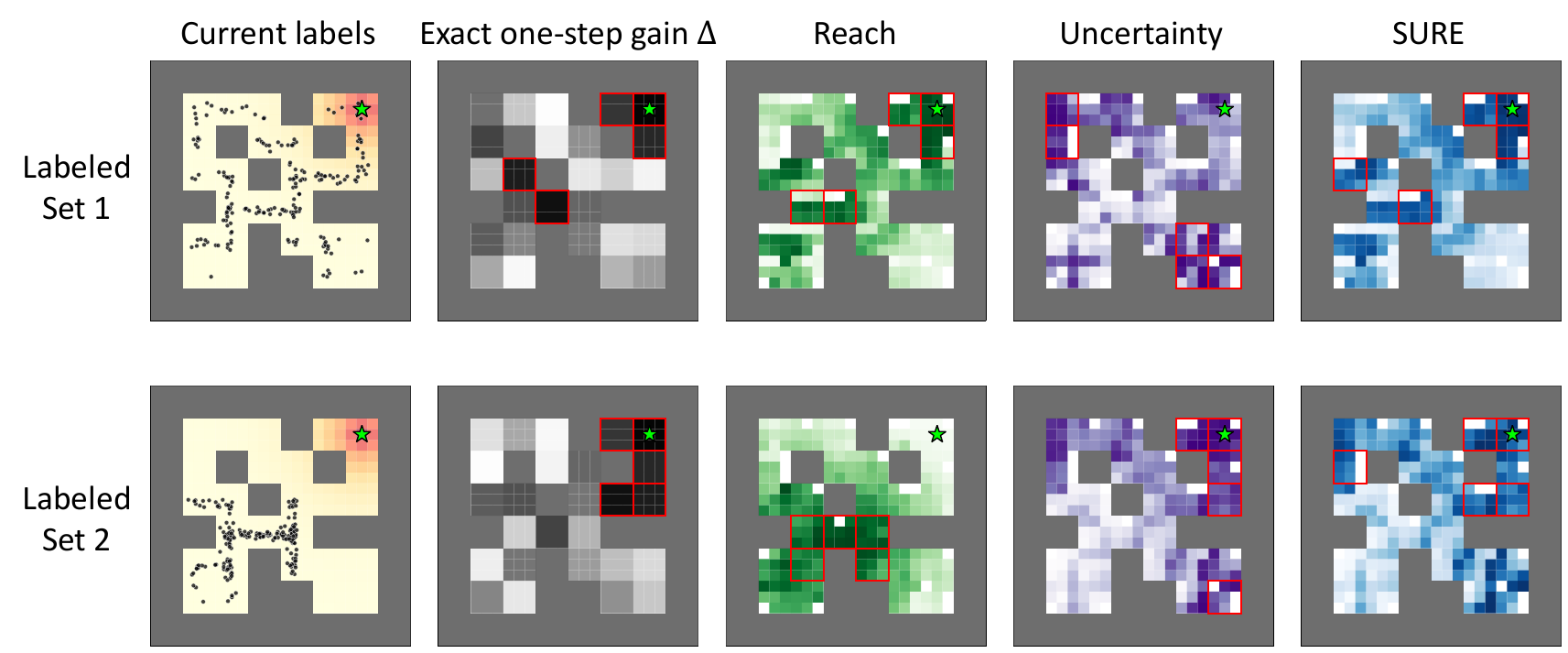}
    \caption{Ablation of the SURE score on representative maze rounds. Each row shows a different stage of selection. From left to right: acquired feedback locations, exact one-step gain $\Delta(s,a;\mathcal Y_{[b]},\rho_b)$, Reach-only score, Uncertainty-only score, and SURE score. Red boxes mark the highest-scoring cells under each criterion, and the green star denotes the goal region.}
    \label{fig:maze_diagnostic}
    \vspace{-5mm}
\end{figure}

Figure~\ref{fig:maze_diagnostic} presents an ablation on the maze domain by visualizing how different acquisition signals rank candidate states under different initial labeled sets. Each row corresponds to one initial labeled set. The first column shows the current labeled samples (black dots) together with the underlying ground-truth reward field (background color), which is not accessible to the selection strategy. The second column shows the \emph{exact} one-step gain, computed by adding additional feedback from each grid cell (sampling 50 transitions from the dataset whose states fall into that cell); darker colors indicate larger gain, meaning selecting from that cell under the current labeled set leads to greater policy improvement. The remaining columns show the selection distributions induced by the Reach score, the Uncertainty score, and the SURE score. Darker colors indicate higher selection density, and the highlighted cells are the top-ranked cells, i.e., the states most likely to be selected by each acquisition rule.

The results show that the alignment between each acquisition rule and the exact one-step gain depends on the structure of the current labeled set, while SURE consistently tracks the high-gain regions. When the labeled set provides broad coverage, the uncertainty field is relatively flat, and the highest-gain cells concentrate along the goal-reaching corridor; in this regime, Reach has stronger overlap with the exact gain, while Uncertainty highlights scattered regions with weaker alignment. Conversely, when the labeled set is spatially biased, the highest-gain cells shift toward previously unexplored areas; here, Uncertainty aligns better with the gain, while Reach concentrates near already-labeled regions. Across both scenarios, SURE overlaps well with the exact-gain cells by combining the two signals rather than committing to either one alone. This distinction becomes more important in higher-dimensional domains, where the labeled set cannot be easily visualized as uniformly spread or clustered, and where the labeled distribution changes as more feedback is acquired. SURE therefore provides a more robust acquisition rule by adapting to both policy relevance and reward uncertainty throughout selection.

\subsection{Robustness to the Offline-RL Backbone}
\label{sec:backbone}

All preceding experiments use the same IQL backbone for every method to isolate differences due to reward-label acquisition. However, SURE is not intrinsically tied to IQL: the analysis in Section~\ref{sec:method} invokes the downstream learner only through the pessimistic offline-learning condition in equation~\eqref{eq:offline_backbone_gap}. To test whether the empirical gains are specific to IQL, we replace the backbone throughout the full acquisition loop, including the policy and value estimates used to construct the SURE score, with CQL~\citep{kumar2020cql} and TD3+BC~\citep{fujimoto2021minimalist} on Maze2d-Medium and HalfCheetah. CQL provides a more theory-aligned test because it explicitly encourages conservative value estimates, whereas TD3+BC tests the acquisition rule with a behavior-regularized learner outside the formal condition.

\begin{table}[t]
\centering
\caption{Percentage feedback required to reach $90\%$ of each backbone's own matched full-feedback reference, averaged over $50$ paired seeds.}
\label{tab:backbone}
\begin{tabular}{llcc}
\toprule
Backbone & Domain & \textsc{SURE} & \textsc{Uniform} \\
\midrule
\multirow{2}{*}{TD3+BC} & Maze2d-Medium & \textbf{0.14\%} & 1.92\% \\
 & HalfCheetah & \textbf{0.52\%} & 2.36\% \\
\midrule
\multirow{2}{*}{CQL} & Maze2d-Medium & \textbf{0.16\%} & 2.00\% \\
 & HalfCheetah & \textbf{0.55\%} & 2.44\% \\
\bottomrule
\end{tabular}
\vspace{-3mm}
\end{table}

Table~\ref{tab:backbone} shows that SURE requires approximately $4.4\times$--$13.7\times$ less feedback than \textsc{Uniform} under both alternative backbones and in both domains, indicating that the acquisition benefit of SURE is not specific to IQL. The sensitivity of SURE to the size of the initial warm-start set is reported in Appendix~\ref{app:warmstart}.

\vspace{-2mm}
\section{Discussion and Conclusion}
\label{sec:conclusion}

This work studies reward selection for RLLF in the retained-label setting, where queried rewards can be reused to train a reward model before downstream offline RL. Our analysis shows that, once the offline dataset is fixed, selection can only improve the reward-model uncertainty term in the suboptimality bound. This clarifies the role of reward uncertainty: uncertainty is most useful to reduce when it lies along directions that good policies are likely to use. We then propose SURE, which turns this principle into a practical acquisition rule by combining successor-guided reachability with uncertainty reduction, so that selected rewards are both connected to promising regions and useful for shrinking the reward confidence radius. The theory justifies this rule by deriving the acquisition objective from the suboptimality bound and showing how computable reference measures can replace the unknown optimal occupancy. Empirically, SURE is consistently more label-efficient across a variety of domains and offline-RL backbones, while other selection strategies are strongest mainly in regimes favored by their respective biases.

\newpage
\begin{ack}
This work used computational resources and services provided by the Unity Research Computing Platform and the Swarm cluster at the University of Massachusetts Amherst. We also thank the anonymous reviewers for their comments and suggestions, which helped strengthen and clarify this work.
\end{ack}

\bibliographystyle{plainnat}
\bibliography{references}

@inproceedings{abbasi2011linear,
  author    = {Abbasi-Yadkori, Yasin and P{\'a}l, D{\'a}vid and Szepesv{\'a}ri, Csaba},
  title     = {Improved Algorithms for Linear Stochastic Bandits},
  booktitle = {Advances in Neural Information Processing Systems},
  pages     = {2312--2320},
  year      = {2011}
}

@article{bai2022constitutional,
  author = {Bai, Yuntao and Kadavath, Saurav and Kundu, Sandipan and Askell, Amanda and Kernion, Jackson and Jones, Andy and Chen, Anna and Goldie, Anna and Mirhoseini, Azalia and McKinnon, Cameron and others},
  title = {Constitutional AI: Harmlessness from AI Feedback},
  journal = {arXiv preprint arXiv:2212.08073},
  year = {2022}
}

@inproceedings{barreto2017successor,
  author = {Barreto, Andr{\'e} and Dabney, Will and Munos, R{\'e}mi and Hunt, Jonathan J. and Schaul, Tom and van Hasselt, Hado and Silver, David},
  title = {Successor Features for Transfer in Reinforcement Learning},
  booktitle = {Advances in Neural Information Processing Systems},
  year = {2017}
}

@article{chaudhari2025which,
  author = {Chaudhari, Shreyas and Zhang, Renhao and Thomas, Philip S. and Castro da Silva, Bruno},
  title = {Which Rewards Matter? Reward Selection for Reinforcement Learning under Limited Feedback},
  journal = {arXiv preprint arXiv:2510.00144},
  year = {2025}
}

@inproceedings{christiano2017preferences,
  author = {Christiano, Paul F. and Leike, Jan and Brown, Tom B. and Martic, Miljan and Legg, Shane and Amodei, Dario},
  title = {Deep Reinforcement Learning from Human Preferences},
  booktitle = {Advances in Neural Information Processing Systems},
  year = {2017}
}

@article{dayan1993successor,
  author = {Dayan, Peter},
  title = {Improving Generalization for Temporal Difference Learning: The Successor Representation},
  journal = {Neural Computation},
  volume = {5},
  number = {4},
  pages = {613--624},
  year = {1993}
}

@inproceedings{hu2023pds,
  author    = {Hu, Hao and Yang, Yiqin and Zhao, Qianchuan and Zhang, Chongjie},
  title     = {The Provable Benefits of Unsupervised Data Sharing for Offline Reinforcement Learning},
  booktitle = {International Conference on Learning Representations},
  year      = {2023}
}

@inproceedings{kostrikov2021iql,
  author    = {Kostrikov, Ilya and Nair, Ashvin and Levine, Sergey},
  title     = {Offline Reinforcement Learning with Implicit Q-Learning},
  booktitle = {International Conference on Learning Representations},
  year      = {2022}
}

@inproceedings{ouyang2022instructgpt,
  author = {Ouyang, Long and Wu, Jeffrey and Jiang, Xu and Almeida, Diogo and Wainwright, Carroll L. and Mishkin, Pamela and Zhang, Chong and Agarwal, Sandhini and Slama, Katarina and Ray, Alex and others},
  title = {Training Language Models to Follow Instructions with Human Feedback},
  booktitle = {Advances in Neural Information Processing Systems},
  year = {2022}
}

@article{lai2025kernelpds,
  author = {Lai, Yen-Ru and Chang, Fu-Chieh and Wu, Pei-Yuan},
  title = {Leveraging Unlabeled Data Sharing through Kernel Function Approximation in Offline Reinforcement Learning},
  journal = {Transactions on Machine Learning Research},
  year = {2025}
}

@inproceedings{parisi2024beyond,
  author = {Parisi, Simone and Kazemipour, Alireza and Bowling, Michael},
  title = {Beyond Optimism: Exploration with Partially Observable Rewards},
  booktitle = {Advances in Neural Information Processing Systems},
  year = {2024}
}

@inproceedings{yu2022unlabeled,
  author = {Yu, Tianhe and Kumar, Aviral and Chebotar, Yevgen and Hausman, Karol and Finn, Chelsea and Levine, Sergey},
  title = {How to Leverage Unlabeled Data in Offline Reinforcement Learning},
  booktitle = {International Conference on Machine Learning},
  year = {2022}
}

@misc{krueger2020active,
  title = {Active Reinforcement Learning: Observing Rewards at a Cost},
  author = {Krueger, David and Leike, Jan and Evans, Owain and Salvatier, John},
  year = {2020},
  eprint = {2011.06709},
  archivePrefix = {arXiv}
}

@inproceedings{sadigh2017active,
  title = {Active Preference-Based Learning of Reward Functions},
  author = {Sadigh, Dorsa and Dragan, Anca D. and Sastry, S. Shankar and Seshia, Sanjit A.},
  booktitle = {Proceedings of Robotics: Science and Systems},
  year = {2017}
}

@inproceedings{biyik2020asking,
  title = {Asking Easy Questions: A User-Friendly Approach to Active Reward Learning},
  author = {B{\i}y{\i}k, Erdem and Palan, Malayandi and Landolfi, Nicholas C. and Losey, Dylan P. and Sadigh, Dorsa},
  booktitle = {Proceedings of the 3rd Conference on Robot Learning},
  year = {2020}
}

@inproceedings{lindner2021information,
  title = {Information Directed Reward Learning for Reinforcement Learning},
  author = {Lindner, David and Turchetta, Matteo and Tschiatschek, Sebastian and Ciosek, Kamil and Krause, Andreas},
  booktitle = {Advances in Neural Information Processing Systems},
  year = {2021}
}

@inproceedings{le2019batch,
  author    = {Le, Hoang and Voloshin, Cameron and Yue, Yisong},
  title     = {Batch Policy Learning under Constraints},
  booktitle = {Proceedings of the 36th International Conference on Machine Learning},
  pages     = {3703--3712},
  year      = {2019}
}

@article{daniel2015active,
  author  = {Daniel, Christian and Kroemer, Oliver and Viering, Malte and Metz, Jan and Peters, Jan},
  title   = {Active Reward Learning with a Novel Acquisition Function},
  journal = {Autonomous Robots},
  volume  = {39},
  number  = {3},
  pages   = {389--405},
  year    = {2015}
}

@inproceedings{wilde2020active,
  title = {Active Preference Learning Using Maximum Regret},
  author = {Wilde, Nils and Kuli{\'c}, Dana and Smith, Stephen L.},
  booktitle = {IEEE/RSJ International Conference on Intelligent Robots and Systems (IROS)},
  year = {2020}
}

@misc{schulze2018active,
  title = {Active Reinforcement Learning with {M}onte-{C}arlo Tree Search},
  author = {Schulze, Sebastian and Evans, Owain},
  year = {2018},
  eprint = {1803.04926},
  archivePrefix = {arXiv}
}

@inproceedings{tucker2023bandits,
  title = {Bandits with Costly Reward Observations},
  author = {Tucker, Aaron David and Biddulph, Caleb and Wang, Claire and Joachims, Thorsten},
  booktitle = {Conference on Uncertainty in Artificial Intelligence},
  year = {2023}
}

@inproceedings{parisi2024monitored,
  title = {Monitored {M}arkov Decision Processes},
  author = {Parisi, Simone and Mohammedalamen, Montaser and Kazemipour, Alireza and Taylor, Matthew E. and Bowling, Michael},
  booktitle = {Proceedings of the International Conference on Autonomous Agents and Multiagent Systems (AAMAS)},
  year = {2024}
}

@misc{romeo2024imputed,
  title = {Offline Reinforcement Learning with Imputed Rewards},
  author = {Romeo, Carlo and Bagdanov, Andrew D.},
  year = {2024},
  eprint = {2407.10839},
  archivePrefix = {arXiv}
}

@misc{konyushkova2020semi,
  title = {Semi-Supervised Reward Learning for Offline Reinforcement Learning},
  author = {Konyushkova, Ksenia and Zolna, Konrad and Aytar, Yusuf and Novikov, Alexander and Reed, Scott and Cabi, Serkan and de Freitas, Nando},
  year = {2020},
  eprint = {2012.06899},
  archivePrefix = {arXiv}
}

@misc{zolna2020offline,
  title = {Offline Learning from Demonstrations and Unlabeled Experience},
  author = {Zolna, Konrad and Novikov, Alexander and Konyushkova, Ksenia and Gulcehre, Caglar and Wang, Ziyu and Aytar, Yusuf and Denil, Misha and de Freitas, Nando and Reed, Scott},
  year = {2020},
  eprint = {2011.13885},
  archivePrefix = {arXiv}
}

@inproceedings{nishimori2025domain,
  title = {Offline Reinforcement Learning with Domain-Unlabeled Data},
  author = {Nishimori, Soichiro and Cai, Xin-Qiang and Ackermann, Johannes and Sugiyama, Masashi},
  booktitle = {Reinforcement Learning Conference},
  year = {2025}
}

@inproceedings{li2023mahalo,
  title = {MAHALO: Unifying Offline Reinforcement Learning and Imitation Learning from Observations},
  author = {Li, Anqi and Boots, Byron and Cheng, Ching-An},
  booktitle = {Proceedings of the 40th International Conference on Machine Learning},
  year = {2023}
}

@inproceedings{jin2021pessimism,
  author    = {Jin, Ying and Yang, Zhuoran and Wang, Zhaoran},
  title     = {Is Pessimism Provably Efficient for Offline Reinforcement Learning?},
  booktitle = {Proceedings of the 38th International Conference on Machine Learning},
  year      = {2021}
}

@inproceedings{kumar2020cql,
  author    = {Kumar, Aviral and Zhou, Aurick and Tucker, George and Levine, Sergey},
  title     = {Conservative Q-Learning for Offline Reinforcement Learning},
  booktitle = {Advances in Neural Information Processing Systems},
  year      = {2020}
}

@inproceedings{jin2020provably,
  author    = {Jin, Chi and Yang, Zhuoran and Wang, Zhaoran and Jordan, Michael I.},
  title     = {Provably Efficient Reinforcement Learning with Linear Function Approximation},
  booktitle = {Conference on Learning Theory (COLT)},
  year      = {2020}
}

@inproceedings{yang2019reinforcement,
  author    = {Yang, Lin and Wang, Mengdi},
  title     = {Reinforcement Learning in Feature Space: Matrix Bandit, Kernels, and Regret Bound},
  booktitle = {Proceedings of the 37th International Conference on Machine Learning},
  pages     = {10746--10756},
  year      = {2020}
}

@book{sutton2018reinforcement,
  author    = {Sutton, Richard S. and Barto, Andrew G.},
  title     = {Reinforcement Learning: An Introduction},
  edition   = {2nd},
  publisher = {MIT Press},
  year      = {2018}
}

@misc{fu2020d4rl,
  author        = {Fu, Justin and Kumar, Aviral and Nachum, Ofir and Tucker, George and Levine, Sergey},
  title         = {{D4RL}: Datasets for Deep Data-Driven Reinforcement Learning},
  year          = {2020},
  eprint        = {2004.07219},
  archivePrefix = {arXiv}
}

@misc{brockman2016gym,
  author        = {Brockman, Greg and Cheung, Vicki and Pettersson, Ludwig and Schneider, Jonas and Schulman, John and Tang, Jie and Zaremba, Wojciech},
  title         = {{OpenAI Gym}},
  year          = {2016},
  eprint        = {1606.01540},
  archivePrefix = {arXiv}
}

@article{reymond2015chemical,
  title={The Chemical Space Project},
  author={Reymond, Jean-Louis},
  journal={Accounts of Chemical Research},
  volume={48},
  number={3},
  pages={722--730},
  year={2015},
  doi={10.1021/ar500432k}
}

@article{gomez2018automatic,
  title={Automatic Chemical Design Using a Data-Driven Continuous Representation of Molecules},
  author={G{\'o}mez-Bombarelli, Rafael and Wei, Jennifer N. and Duvenaud, David and Hern{\'a}ndez-Lobato, Jos{\'e} Miguel and S{\'a}nchez-Lengeling, Benjam{\'i}n and Sheberla, Dennis and Aguilera-Iparraguirre, Jorge and Hirzel, Timothy D. and Adams, Ryan P. and Aspuru-Guzik, Al{\'a}n},
  journal={ACS Central Science},
  volume={4},
  number={2},
  pages={268--276},
  year={2018},
  doi={10.1021/acscentsci.7b00572}
}

@inproceedings{jin2018junction,
  title={Junction Tree Variational Autoencoder for Molecular Graph Generation},
  author={Jin, Wengong and Barzilay, Regina and Jaakkola, Tommi},
  booktitle={Proceedings of the 35th International Conference on Machine Learning},
  pages={2323--2332},
  year={2018}
}

@article{popova2018deep,
  title={Deep Reinforcement Learning for de novo Drug Design},
  author={Popova, Mariya and Isayev, Olexandr and Tropsha, Alexander},
  journal={Science Advances},
  volume={4},
  number={7},
  pages={eaap7885},
  year={2018},
  doi={10.1126/sciadv.aap7885}
}

@article{dimasi2016innovation,
  title={Innovation in the Pharmaceutical Industry: New Estimates of {R\&D} Costs},
  author={DiMasi, Joseph A. and Grabowski, Henry G. and Hansen, Ronald W.},
  journal={Journal of Health Economics},
  volume={47},
  pages={20--33},
  year={2016},
  doi={10.1016/j.jhealeco.2016.01.012}
}

@inproceedings{fujimoto2021minimalist,
  author    = {Fujimoto, Scott and Gu, Shixiang Shane},
  title     = {A Minimalist Approach to Offline Reinforcement Learning},
  booktitle = {Advances in Neural Information Processing Systems},
  year      = {2021}
}

@article{chaudhari2026data,
  author    = {Chaudhari, Shreyas and Zhang, Renhao and Thomas, Philip S. and Castro da Silva, Bruno},
  title     = {Data Selection for Reward Labeling in Limited-Feedback Reinforcement Learning},
  journal   = {Transactions on Machine Learning Research},
  issn      = {2835-8856},
  year      = {2026},
  url       = {https://openreview.net/forum?id=N8OlHGeHSy}
}

@inproceedings{zhang2024model,
  author    = {Zhang, Renhao and Fu, Haotian and Miao, Yilin and Konidaris, George},
  title     = {Model-based Reinforcement Learning for Parameterized Action Spaces},
  booktitle = {International Conference on Machine Learning},
  year      = {2024}
}

@inproceedings{zhang2026noise,
  author    = {Zhang, Renhao and Fu, Haotian and Jia, Mingxi and Konidaris, George and Du, Yilun and Castro da Silva, Bruno},
  title     = {From Noise to Control: Parameterized Diffusion Policies},
  booktitle = {International Conference on Machine Learning},
  year      = {2026}
}

@inproceedings{zhang2026evidence,
  author    = {Zhang, Renhao and Li, Mingzhe and Li, Yuli and Zheng, Jiayu and Miao, Yilin},
  title     = {Learning What to Keep: Evidence-Gated Parameterized Skills for Frozen {LLM} Agents},
  booktitle = {NeurIPS 2026 Workshop on Test-Time Continual Learning},
  year      = {2026}
}

@inproceedings{zhang2026latent,
  author    = {Zhang, Renhao and Li, Mingzhe and Castro da Silva, Bruno},
  title     = {Latent Barrier Steering: Hierarchical Safety for Generative Planning},
  booktitle = {Advances in Neural Information Processing Systems},
  year      = {2026}
}

\newpage
\appendix

\section{\SURE{} Algorithm}
\label{app:algorithm}

Algorithm~\ref{alg:sgpds} gives the full pseudocode for \SURE{}, assembled from the \SURE{} score~\eqref{eq:swvr}, the reachability measure~\eqref{eq:mixed_measure}, the successor-guided distribution \(\bar H_b\) of Appendix~\ref{app:normalized_reachability} obtained by normalizing the per-candidate scores~\eqref{eq:reachability_score}, and the pessimistic reward and backbone objects of Section~\ref{sec:prelim}. Section~\ref{sec:method} uses single-query notation to simplify the presentation of the acquisition rule and its analysis; in the experiments, labels are acquired in batches. \SURE{} starts from \(B_0\) uniformly sampled warm-start labels, which are shared by all methods within each seed and counted toward the feedback budget \(B\). Each subsequent round \(b\) acquires the \(M\) unlabeled candidates with the largest \SURE{} scores, after which the reward model, the offline backbone, the successor features, and the acquisition scores are recomputed. With \(B_0=0\) and \(M=1\), round \(b\) coincides with the \(b\)-th query and Algorithm~\ref{alg:sgpds} reduces to the single-query rule of Section~\ref{sec:method}. We use \(B_0=M=64\) in all continuous-control experiments (Appendix~\ref{app:hyperparams}). After the budget is exhausted, \SURE{} refits the reward model using the final labeled set \(\mathcal Y^{[R]}\), re-runs the pessimistic backbone once more on the full dataset, and returns the resulting policy \(\pi^{[R]}\).

\begin{algorithm}[h]
\caption{\SURE{}: Successor-Guided Uncertainty REduction for reward labeling}
\label{alg:sgpds}
\begin{algorithmic}[1]
\Require offline dataset \(\mathcal D\); budget \(B\); warm-start size \(B_0\); batch size \(M\); top-\(K\); weights \(\{\beta_b\}\).
\State Draw \(\mathcal S_0\subset\mathcal D\) with \(|\mathcal S_0|=B_0\) uniformly at random; query their rewards to form \(\mathcal Y^{[0]}\).
\State Set the number of acquisition rounds \(R\gets (B-B_0)/M\).
\For{\(b=0,1,\ldots,R-1\)}
  \State Fit ridge reward estimator \(\hat\theta_b\) and build pessimistic reward \(\tilde r_b\) from \(\mathcal Y^{[b]}\).
  \State Run the offline backbone on \(\mathcal D\) with reward \(\tilde r_b\) to obtain \(\pi^{[b]}\) and \(Q_b\).
  \State Form target set \(T_b\subset \mathcal D\): the top-\(K\) transitions in \(\mathcal D\) under \(Q_b\).
  \State Compute per-candidate scores \(H_b(s_i,a_i)\) via \eqref{eq:reachability_score} and the successor-guided distribution \(\bar H_b\) via Appendix~\ref{app:normalized_reachability}.
  \State Mix with behavior: \(\rho_b=\beta_b\hat d_{\mathrm{beh}}^{(b)}+(1-\beta_b)\bar H_b\).
  \State \(U_b\gets\) the \(M\) candidates in \(\mathcal D\setminus\mathcal S_b\) with the largest \(\SUREscore_b(s_i,a_i)\) in \eqref{eq:swvr}.
  \State Query \(r_i=r(s_i,a_i)\) for all \((s_i,a_i,s_i')\in U_b\); update \(\mathcal Y^{[b+1]}\gets \mathcal Y^{[b]}\cup\{(s_i,a_i,r_i,s_i')\}_{(s_i,a_i,s_i')\in U_b}\), \(\mathcal S_{b+1}\gets \mathcal S_b\cup U_b\).
\EndFor
\State Refit \(\tilde r_R\) from \(\mathcal Y^{[R]}\); run backbone on \(\mathcal D\) with \(\tilde r_R\) to obtain \(\pi^{[R]}\).
\State \Return \(\pi^{[R]}\).
\end{algorithmic}
\end{algorithm}

\section{Derivations for Section~\ref{sec:prelim}}
\label{app:prelim_derivations}

This appendix collects the derivations of the constructions used in Section~\ref{sec:prelim}: the PDS pipeline that produces the pessimistic reward estimate together with the reward-confidence bound~\eqref{eq:pessimistic_reward_gap}, a ridge estimator for the one-step feature transition operator that supports the successor features of Definition~\ref{def:successor_feature}, and the offline-RL error term \(\Gamma_{\rm off}(\mathcal D,\delta)\) used in equation~\eqref{eq:offline_backbone_gap}.

\subsection{PDS pipeline and reward-confidence bound}
\label{app:reward_confidence}
\label{app:pds_pipeline}

The ``Reward Modeling for Partially Labeled Offline RL'' paragraph of Section~\ref{sec:prelim} describes how the retained labeled set \(\mathcal Y\) is converted into a reward signal that the downstream offline-RL backbone can consume. We restate that pipeline here in pragmatic terms, formalize the ridge regression step, derive the reward-confidence bound~\eqref{eq:pessimistic_reward_gap}, and verify that the resulting \(\tilde r_{\mathcal Y}\) underestimates the true reward at every state--action pair.

\paragraph{The PDS pipeline.}
The pipeline composes two ingredients. First, ridge regression on the labeled set \(\mathcal Y\) (equation~\eqref{eq:reward_ridge_objective}) fits a single linear reward parameter \(\hat\theta_{\mathcal Y}\). Because the reward model is linear in \(\phi(s,a)\), this single estimator produces a prediction \(\phi(s,a)^\top\hat\theta_{\mathcal Y}\) at every state--action pair---not only the queried ones---which is what allows a small number of labels to generate a reward signal everywhere the offline pool reaches. Second, PDS turns this point prediction into a lower-confidence reward
\[
\tilde r_{\mathcal Y}(s,a)
=
\bigl[\phi(s,a)^\top\hat\theta_{\mathcal Y}-\alpha_{\mathcal Y}\sigma_{\mathcal Y}(s,a)\bigr]_+
\]
by subtracting the elliptical uncertainty radius \(\sigma_{\mathcal Y}(s,a)\) from the ridge prediction, scaled by the width \(\alpha_{\mathcal Y}\). By construction, the result underestimates the true reward, \(\tilde r_{\mathcal Y}\le r\) with high probability, so the downstream pessimistic backbone (e.g., PEVI~\citep{jin2021pessimism}) inherits a clean reward-side guarantee.

\paragraph{Matrix-form derivation of the ridge estimator.}
Fix a labeled set \(\mathcal Y\subseteq\mathcal D\), and let \(n:=|\mathcal Y|\). Enumerate it as
\(
\mathcal Y=\{(s_1,a_1,y_1),\ldots,(s_n,a_n,y_n)\}.
\)
Define the design matrix, the noise vector, and the response vector
\[
X
:=
\begin{bmatrix}
\phi(s_1,a_1)^\top\\
\vdots\\
\phi(s_n,a_n)^\top
\end{bmatrix}
\in\mathbb R^{n\times d},
\qquad
\xi:=(\xi_1,\ldots,\xi_n)^\top\in\mathbb R^{n},
\qquad
y:=(y_1,\ldots,y_n)^\top.
\]
Under the linear reward model, \(y=X\theta^\star+\xi\). In matrix form, the ridge objective~\eqref{eq:reward_ridge_objective} reads \(L(\theta)=\|y-X\theta\|_2^2+\lambda\|\theta\|_2^2\). Setting its gradient to zero gives the normal equation
\[
\Lambda(\mathcal Y)=\lambda I+X^\top X,
\qquad
\hat\theta_{\mathcal Y}=\Lambda(\mathcal Y)^{-1}X^\top y.
\]
Substituting \(y=X\theta^\star+\xi\) and using \(X^\top X-\Lambda(\mathcal Y)=-\lambda I\), we obtain
\begin{equation}
\label{eq:ridge_error_decomp}
\hat\theta_{\mathcal Y}-\theta^\star
=
\Lambda(\mathcal Y)^{-1}X^\top\xi
-
\lambda\Lambda(\mathcal Y)^{-1}\theta^\star.
\end{equation}
The first term is the stochastic estimation error inherited from the noise \(\xi\), and the second is the deterministic ridge bias toward zero induced by \(\lambda\). Bounding each yields the reward-confidence inequality below.

\paragraph{Derivation of the reward-confidence bound.}

\begin{proof}[Derivation of~\eqref{eq:pessimistic_reward_gap}]
We bound the two terms of \eqref{eq:ridge_error_decomp} separately, combine them, and apply the nonnegative-reward clipping of PDS.

\emph{Step 1: bounding the stochastic term.}
Fix \((s,a)\in\mathcal S\times\mathcal A\). Using \eqref{eq:ridge_error_decomp},
\[
\phi(s,a)^\top(\hat\theta_{\mathcal Y}-\theta^\star)
=
\phi(s,a)^\top\Lambda(\mathcal Y)^{-1}X^\top\xi
-
\lambda\phi(s,a)^\top\Lambda(\mathcal Y)^{-1}\theta^\star.
\]
For the first term, insert \(\Lambda(\mathcal Y)^{-1/2}\Lambda(\mathcal Y)^{1/2}\) and apply Cauchy--Schwarz:
\[
\bigl|\phi(s,a)^\top\Lambda(\mathcal Y)^{-1}X^\top\xi\bigr|
\le
\|\Lambda(\mathcal Y)^{-1/2}\phi(s,a)\|_2\cdot\|\Lambda(\mathcal Y)^{-1/2}X^\top\xi\|_2
=
\sigma_{\mathcal Y}(s,a)\cdot\|X^\top\xi\|_{\Lambda(\mathcal Y)^{-1}},
\]
using \(\|\Lambda(\mathcal Y)^{-1/2}\phi(s,a)\|_2=\sigma_{\mathcal Y}(s,a)\). By the self-normalized concentration inequality of \citet{abbasi2011linear} for linear regression with conditionally \(\sigma_r\)-sub-Gaussian noise, with probability at least \(1-\delta\), uniformly over all labeled sets \(\mathcal Y\subseteq\mathcal D\) the learner may encounter,
\[
\|X^\top\xi\|_{\Lambda(\mathcal Y)^{-1}}
\le
\sigma_r\sqrt{2\log\!\Bigl(\tfrac{\det(\Lambda(\mathcal Y))^{1/2}}{\det(\lambda I)^{1/2}\delta}\Bigr)}.
\]
Bounding the determinant ratio by the AM--GM inequality together with \(\sum_j\eta_j=\tr(X^\top X)\le n\) (since \(\|\phi(s,a)\|_2\le 1\)) gives \(\det(\Lambda(\mathcal Y))/\det(\lambda I)\le(1+n/\lambda)^d\), hence
\begin{equation}
\label{eq:stochastic_term_final}
\bigl|\phi(s,a)^\top\Lambda(\mathcal Y)^{-1}X^\top\xi\bigr|
\le
\sigma_{\mathcal Y}(s,a)\,\sigma_r\sqrt{2\log(1/\delta)+d\log(1+n/\lambda)}.
\end{equation}

\emph{Step 2: bounding the ridge-bias term.}
Insert \(\Lambda(\mathcal Y)^{-1/2}\Lambda(\mathcal Y)^{1/2}\) again and apply Cauchy--Schwarz:
\[
\bigl|\lambda\phi(s,a)^\top\Lambda(\mathcal Y)^{-1}\theta^\star\bigr|
\le
\sqrt{\lambda}\,\|\Lambda(\mathcal Y)^{-1/2}\phi(s,a)\|_2
\cdot
\sqrt{\lambda}\,\|\Lambda(\mathcal Y)^{-1/2}\theta^\star\|_2
\le
\sqrt{\lambda}\,S_r\,\sigma_{\mathcal Y}(s,a),
\]
where the last step uses \(\Lambda(\mathcal Y)\succeq\lambda I\), so \(\sqrt{\lambda}\|\Lambda(\mathcal Y)^{-1/2}\|_{\rm op}\le 1\), and \(\|\theta^\star\|_2\le S_r\).

\emph{Step 3: combining and clipping.}
Adding the two bounds,
\[
\bigl|\phi(s,a)^\top(\hat\theta_{\mathcal Y}-\theta^\star)\bigr|
\le
\bigl[\sqrt{\lambda}\,S_r+\sigma_r\sqrt{2\log(1/\delta)+d\log(1+n/\lambda)}\bigr]\sigma_{\mathcal Y}(s,a)
=
\alpha_{\mathcal Y}\sigma_{\mathcal Y}(s,a).
\]
Define \(\hat r_{\mathcal Y}(s,a):=\phi(s,a)^\top\hat\theta_{\mathcal Y}\), so \(|r(s,a)-\hat r_{\mathcal Y}(s,a)|\le\alpha_{\mathcal Y}\sigma_{\mathcal Y}(s,a)\). Recall \(\tilde r_{\mathcal Y}(s,a)=\max\{\hat r_{\mathcal Y}(s,a)-\alpha_{\mathcal Y}\sigma_{\mathcal Y}(s,a),0\}\). When the un-clipped expression is nonnegative, the displayed two-sided confidence bound gives \(0\le r(s,a)-\tilde r_{\mathcal Y}(s,a)\le 2\alpha_{\mathcal Y}\sigma_{\mathcal Y}(s,a)\); when it is negative, \(\tilde r_{\mathcal Y}(s,a)=0\) and \(r(s,a)\le\hat r_{\mathcal Y}(s,a)+\alpha_{\mathcal Y}\sigma_{\mathcal Y}(s,a)<2\alpha_{\mathcal Y}\sigma_{\mathcal Y}(s,a)\), again yielding the same two-sided gap. This is equation~\eqref{eq:pessimistic_reward_gap}.
\end{proof}

\subsection{Successor-feature estimation}
\label{app:m_pi}

Under Definition~\ref{def:linear_mdp}, the successor feature of Definition~\ref{def:successor_feature} admits the closed form
\[
\psi^\pi(s,a)=(I-\gamma M_\pi)^{-1}\phi(s,a),
\qquad
M_\pi\in\mathbb R^{d\times d},
\]
where \(M_\pi\) is the one-step feature transition operator under \(\pi\), characterized by
\(
M_\pi\phi(s,a)=\E^\pi[\phi(s',a')\mid (s,a)]
\)
with \(s'\sim p(\cdot\mid s,a)\) and \(a'\sim\pi(\cdot\mid s')\). Given the offline transition dataset \(\mathcal D=\{(s_t,a_t,s'_t)\}\) and a policy \(\hat\pi\), we estimate \(M_{\hat\pi}\) by ridge regression of the next-feature expectation on the current feature:
\begin{equation}
\label{eq:m_pi_ridge}
\hat M_{\hat\pi}
=
\arg\min_{M\in\mathbb R^{d\times d}}
\sum_{(s,a,s')\in\mathcal D}
\bigl\|\E_{a'\sim\hat\pi(\cdot\mid s')}[\phi(s',a')]-M\phi(s,a)\bigr\|_2^2
+
\lambda_{\rm SF}\|M\|_F^2,
\end{equation}
whose closed form is the standard multivariate ridge estimator. We then set
\(
\hat\psi^{\hat\pi}(s,a)=(I-\gamma\hat M_{\hat\pi})^{-1}\phi(s,a)
\)
and use \(\hat\psi^{\hat\pi}\) in place of \(\psi^{\hat\pi}\) in the per-candidate scores \(H_b\) of equation~\eqref{eq:reachability_score}. The estimation error \(\|\hat M_{\hat\pi}-M_{\hat\pi}\|\) contributes an additive \(\mathcal O(\varepsilon_{\rm SF})\) term to the unified design bound of Appendix~\ref{app:coverage_transfer}; because this term does not depend on the labeled subset \(\mathcal Y\), it does not affect the query-complexity structure of the analysis.

\subsection{Form of the offline-RL error term}
\label{app:gamma_off_form}

The backbone guarantee~\eqref{eq:offline_backbone_gap} is invoked abstractly in the analysis. Here we record the concrete instantiation we use in experiments---the pessimistic value-iteration (PEVI) backbone of \citet{jin2021pessimism} adapted to retained-reward partial labels by \citet{hu2023pds}---and derive the \(\max\)-over-\(\mathcal D\) form of \(\Gamma_{\rm off}(\mathcal D,\delta)\) used in Section~\ref{sec:prelim}.

Let \(\Lambda(\mathcal D):=\lambda_{\rm off}I+\sum_{i=1}^{N}\phi(s_i,a_i)\phi(s_i,a_i)^\top\) be the offline value-regression design matrix built from \emph{all} transitions in \(\mathcal D\), with regularizer \(\lambda_{\rm off}\ge 1\). The associated offline pessimism radius is \(\sigma_{\rm off}(s,a):=\sqrt{\phi(s,a)^\top\Lambda(\mathcal D)^{-1}\phi(s,a)}\), and the offline confidence width is
\[
\beta_{\rm off}
:=
c_{\rm off}\,
\Bigl(
\sqrt{\lambda_{\rm off}}\,(1-\gamma)^{-1}r_{\max}
+
\sqrt{d\log(N/\delta)}\,(1-\gamma)^{-1}
\Bigr),
\]
for an absolute constant \(c_{\rm off}>0\) inherited from the self-normalized vector concentration of \citet{abbasi2011linear}. By construction, \(\beta_{\rm off}\), \(\sigma_{\rm off}\), and \(\Lambda(\mathcal D)\) are functions of \((\mathcal D,\delta)\) only.

Theorem~4.4 of \citet{jin2021pessimism}, instantiated in the linear-MDP setting of Definition~\ref{def:linear_mdp} and combined with the pessimistic-reward clipping of \citet{hu2023pds}, yields, with probability at least \(1-\delta\), the PEVI suboptimality bound
\begin{equation}
\label{eq:pevi_raw_bound}
J_{\tilde r}(\pi^\star_{\tilde r})-J_{\tilde r}(\hat\pi_{\tilde r})
\le
\frac{2\beta_{\rm off}}{1-\gamma}\,
\E_{(s,a)\sim d^{\pi^\star_{\tilde r}}}\!\bigl[\sigma_{\rm off}(s,a)\bigr].
\end{equation}
The right-hand side has a residual \(\tilde r\)-dependence through the integration measure \(d^{\pi^\star_{\tilde r}}\), but the integrand \(\sigma_{\rm off}(s,a)\) is reward- and labeling-free: a function of \((\mathcal D,\delta)\) only. We absorb the residual dependence by replacing the expectation with the maximum of the integrand over the transitions in \(\mathcal D\). Pessimism in PEVI prevents the learned policy from leaving the support of \(\mathcal D\), so \(d^{\pi^\star_{\tilde r}}\) effectively concentrates on the state--action pairs \(\mathcal D\) covers, and the chain
\begin{equation}
\label{eq:gamma_off_form}
J_{\tilde r}(\pi^\star_{\tilde r})-J_{\tilde r}(\hat\pi_{\tilde r})
\le
\frac{2\beta_{\rm off}}{1-\gamma}\,
\E_{(s,a)\sim d^{\pi^\star_{\tilde r}}}\!\bigl[\sigma_{\rm off}(s,a)\bigr]
\le
\frac{2\beta_{\rm off}}{1-\gamma}\,
\max_{(s,a)\in\mathcal D}\sigma_{\rm off}(s,a)
=:
\Gamma_{\rm off}(\mathcal D,\delta)
\end{equation}
holds. The last equality is the definition of \(\Gamma_{\rm off}(\mathcal D,\delta)\) used in Section~\ref{sec:prelim}: by construction it is a function of \((\mathcal D,\delta)\) alone, with no dependence on which transitions are reward-labeled. The full PEVI / PDS derivation of equation~\eqref{eq:pevi_raw_bound} is given by \citet{jin2021pessimism,hu2023pds}; we do not reproduce it here.

\section{Proofs and Auxiliary Results for Section~\ref{sec:method}}
\label{app:method_proofs}

This appendix proves the formal statements introduced in Section~\ref{sec:method} and supplies the auxiliary constructions deferred from the main text. Throughout, all proofs are carried out on the intersection of the reward-confidence event of \eqref{eq:pessimistic_reward_gap} and the success event of the pessimistic offline-RL backbone from \eqref{eq:offline_backbone_gap}. On this event, all displayed inequalities are deterministic.

\subsection{Proof of Theorem~\ref{thm:label_sensitive}}
\label{app:proof_label_sensitive}

\begin{theorem*}[Label-sensitive suboptimality decomposition]
Under Definition~\ref{def:linear_mdp}, equation~\eqref{eq:pessimistic_reward_gap}, and equation~\eqref{eq:offline_backbone_gap}, with probability at least \(1-2\delta\), for every round \(b\in\{0,\ldots,B\}\),
\[
\SubOpt(\mathcal Q^{(b)})
\le
\Gamma_{\mathrm{off}}(\mathcal D,\delta)
+
\frac{2\alpha_b}{1-\gamma}
\sqrt{\E_{(s,a)\sim d^{\pi_r^\star}}[\sigma_b(s,a)^2]}.
\]
\end{theorem*}

\begin{proof}
Write \(\pi^{[b]}:=\hat\pi_{\tilde r_b}\) for the policy produced by the pessimistic offline-RL backbone in round \(b\). By the definition of suboptimality in Theorem~\ref{thm:label_sensitive},
\[
\SubOpt(\mathcal Q^{(b)})
=
J_r(\pi_r^\star)-J_r(\pi^{[b]}).
\]
Add and subtract two terms involving the pessimistic reward \(\tilde r_b\):
\[
J_r(\pi_r^\star)-J_r(\pi^{[b]})
=
\underbrace{\bigl(J_r(\pi_r^\star)-J_{\tilde r_b}(\pi_r^\star)\bigr)}_{(I)}
+
\underbrace{\bigl(J_{\tilde r_b}(\pi_r^\star)-J_{\tilde r_b}(\pi^{[b]})\bigr)}_{(II)}
+
\underbrace{\bigl(J_{\tilde r_b}(\pi^{[b]})-J_r(\pi^{[b]})\bigr)}_{(III)}.
\]
We bound the three terms separately.

\paragraph{Step 1: bounding term \((III)\).}
Equation~\eqref{eq:pessimistic_reward_gap} gives \(\tilde r_b(s,a)\le r(s,a)\) for every \((s,a)\), and the discounted return is monotone in the reward, so \(J_{\tilde r_b}(\pi)\le J_r(\pi)\) for any policy \(\pi\). Applied to \(\pi=\pi^{[b]}\),
\[
(III)=J_{\tilde r_b}(\pi^{[b]})-J_r(\pi^{[b]})\le 0.
\]

\paragraph{Step 2: bounding term \((II)\).}
Let \(\pi^\star_{\tilde r_b}\in\arg\max_\pi J_{\tilde r_b}(\pi)\) be an optimal policy for the pessimistic reward. Since \(\pi^\star_{\tilde r_b}\) is optimal for \(\tilde r_b\), \(J_{\tilde r_b}(\pi_r^\star)\le J_{\tilde r_b}(\pi^\star_{\tilde r_b})\), and therefore
\[
(II)
\le
J_{\tilde r_b}(\pi^\star_{\tilde r_b})-J_{\tilde r_b}(\pi^{[b]}).
\]
With \(\pi^{[b]}=\hat\pi_{\tilde r_b}\), the backbone guarantee in equation~\eqref{eq:offline_backbone_gap} gives
\[
J_{\tilde r_b}(\pi^\star_{\tilde r_b})-J_{\tilde r_b}(\pi^{[b]})
\le
\Gamma_{\mathrm{off}}(\mathcal D,\delta),
\]
and hence \((II)\le\Gamma_{\mathrm{off}}(\mathcal D,\delta)\).

\paragraph{Step 3: bounding term \((I)\).}
By definition of value,
\[
(I)
=
\E_\eta^{\pi_r^\star}\!\left[
\sum_{t=0}^\infty\gamma^t\bigl(r(s_t,a_t)-\tilde r_b(s_t,a_t)\bigr)
\right].
\]
Equation~\eqref{eq:pessimistic_reward_gap} gives \(0\le r(s_t,a_t)-\tilde r_b(s_t,a_t)\le 2\alpha_b\sigma_b(s_t,a_t)\) for every \(t\), so
\[
(I)
\le
2\alpha_b\,
\E_\eta^{\pi_r^\star}\!\left[
\sum_{t=0}^\infty\gamma^t\sigma_b(s_t,a_t)
\right].
\]
The discounted occupancy identity \(\E_\eta^{\pi_r^\star}\bigl[\sum_t\gamma^t f(s_t,a_t)\bigr]=(1-\gamma)^{-1}\E_{(s,a)\sim d^{\pi_r^\star}}[f(s,a)]\), applied with \(f(s,a)=\sigma_b(s,a)\), yields
\[
(I)
\le
\frac{2\alpha_b}{1-\gamma}\,
\E_{(s,a)\sim d^{\pi_r^\star}}[\sigma_b(s,a)].
\]

\paragraph{Step 4: assembling the pieces.}
Combining the three bounds gives
\[
\SubOpt(\mathcal Q^{(b)})
\le
\Gamma_{\mathrm{off}}(\mathcal D,\delta)
+
\frac{2\alpha_b}{1-\gamma}\,
\E_{(s,a)\sim d^{\pi_r^\star}}[\sigma_b(s,a)].
\]
Since \(\sigma_b(s,a)\ge 0\) and \(d^{\pi_r^\star}\) is a probability measure, Jensen's inequality applied to the concave square-root yields
\(
\E_{(s,a)\sim d^{\pi_r^\star}}[\sigma_b(s,a)]\le\sqrt{\E_{(s,a)\sim d^{\pi_r^\star}}[\sigma_b(s,a)^2]},
\)
and substitution gives the claimed bound.
\end{proof}

\subsection{Proof of Proposition~\ref{prop:exact_gain}}
\label{app:proof_exact_gain}

\begin{proposition*}[Exact one-step marginal gain]
For any reference measure \(\mu\) over the pool and any candidate \((s,a)\),
\[
\Delta(s,a;\mathcal Y^{[b]},\mu)
=
\Phi(\mathcal Y^{[b]};\mu)-\Phi(\mathcal Y^{[b]}\cup\{(s,a)\};\mu)
=
\frac{\|\Sigma(\mu)^{1/2}\Lambda_b^{-1}\phi(s,a)\|_2^2}{1+\sigma_b(s,a)^2}.
\]
In particular, \(\Phi(\mathcal Y^{[b]};\mu)\) is monotonically nonincreasing in the labeled set.
\end{proposition*}

\begin{proof}
For readability, write \(\Lambda:=\Lambda_b\), \(\Sigma:=\Sigma(\mu)\), \(\phi:=\phi(s,a)\), and \(\Lambda^+:=\Lambda+\phi\phi^\top\). The design potential admits a trace form: by definition \(\sigma_b(s,a)^2=\phi(s,a)^\top\Lambda^{-1}\phi(s,a)=\tr(\Lambda^{-1}\phi(s,a)\phi(s,a)^\top)\), and taking expectation over \((s,a)\sim\mu\) with linearity of trace gives
\begin{equation}
\label{eq:phi_trace_identity}
\Phi(\mathcal Y^{[b]};\mu)
=
\E_{(s,a)\sim\mu}[\sigma_b(s,a)^2]
=
\tr\!\bigl(\Sigma(\mu)\Lambda_b^{-1}\bigr).
\end{equation}
Applying \eqref{eq:phi_trace_identity} to both \(\mathcal Y^{[b]}\) and \(\mathcal Y^{[b]}\cup\{(s,a)\}\),
\[
\Phi(\mathcal Y^{[b]};\mu)=\tr(\Sigma\Lambda^{-1}),
\qquad
\Phi(\mathcal Y^{[b]}\cup\{(s,a)\};\mu)=\tr(\Sigma(\Lambda^+)^{-1}),
\]
so
\begin{equation}
\label{eq:delta_start}
\Delta(s,a;\mathcal Y^{[b]},\mu)
=
\tr(\Sigma\Lambda^{-1})-\tr(\Sigma(\Lambda^+)^{-1}).
\end{equation}

\paragraph{Step 1: Sherman--Morrison update.}
Since \(\Lambda\) is positive definite, the rank-one inverse formula applies:
\[
(\Lambda+\phi\phi^\top)^{-1}
=
\Lambda^{-1}
-
\frac{\Lambda^{-1}\phi\phi^\top\Lambda^{-1}}{1+\phi^\top\Lambda^{-1}\phi}.
\]
Substituting into \eqref{eq:delta_start} and using linearity of trace, the \(\tr(\Sigma\Lambda^{-1})\) terms cancel, leaving
\[
\Delta(s,a;\mathcal Y^{[b]},\mu)
=
\frac{\tr\!\bigl(\Sigma\Lambda^{-1}\phi\phi^\top\Lambda^{-1}\bigr)}{1+\phi^\top\Lambda^{-1}\phi}.
\]

\paragraph{Step 2: simplifying the denominator.}
By definition of the uncertainty radius, \(\phi^\top\Lambda^{-1}\phi=\sigma_b(s,a)^2\), so the denominator is \(1+\sigma_b(s,a)^2\).

\paragraph{Step 3: simplifying the numerator.}
By cyclicity of trace and the existence of a symmetric positive semidefinite square root \(\Sigma^{1/2}\),
\[
\tr\!\bigl(\Sigma\Lambda^{-1}\phi\phi^\top\Lambda^{-1}\bigr)
=
\phi^\top\Lambda^{-1}\Sigma\Lambda^{-1}\phi
=
\|\Sigma^{1/2}\Lambda^{-1}\phi\|_2^2.
\]
Substituting yields
\[
\Delta(s,a;\mathcal Y^{[b]},\mu)
=
\frac{\|\Sigma(\mu)^{1/2}\Lambda_b^{-1}\phi(s,a)\|_2^2}{1+\sigma_b(s,a)^2},
\]
the claimed formula.

\paragraph{Step 4: monotonicity.}
The numerator is a squared Euclidean norm and therefore nonnegative, and \(1+\sigma_b(s,a)^2\ge 1>0\), so \(\Delta(s,a;\mathcal Y^{[b]},\mu)\ge 0\) for every \((s,a)\). By definition of \(\Delta\),
\(
\Phi(\mathcal Y^{[b]}\cup\{(s,a)\};\mu)
=
\Phi(\mathcal Y^{[b]};\mu)-\Delta(s,a;\mathcal Y^{[b]},\mu)
\le
\Phi(\mathcal Y^{[b]};\mu),
\)
so \(\Phi\) is monotonically nonincreasing in the labeled set.
\end{proof}

\subsection{Marginal gain and bound decrease}
\label{app:delta_monotonicity}

The bound in equation~\eqref{eq:method_phi_bound} is monotone in \(\Phi(\mathcal Y^{[b]};\mu)\) for any fixed reference measure \(\mu\), so the largest one-step decrease of the bound is achieved by selecting the candidate that maximizes the marginal gain \(\Delta(\cdot;\mathcal Y^{[b]},\mu)\). We make this precise.

For a fixed reference measure \(\mu\), define the bound-side objective
\begin{equation}
\label{eq:def_bound_objective}
U_b(\mu)
:=
\Gamma_{\mathrm{off}}(\mathcal D,\delta)
+
\frac{2\alpha_b}{1-\gamma}\sqrt{\Phi(\mathcal Y^{[b]};\mu)},
\end{equation}
and let \(U_b^{(s,a)}(\mu)\) denote the same quantity after adding \((s,a)\) to the labeled set.

\begin{lemma}[Marginal gain orders bound improvement]
\label{lem:delta_monotone_bound}
Fix a reference measure \(\mu\) and round \(b\). For any candidate \((s,a)\) with \(\Delta(s,a;\mathcal Y^{[b]},\mu)\le\Phi(\mathcal Y^{[b]};\mu)\),
\begin{equation}
\label{eq:bound_decrease_delta}
U_b(\mu)-U_b^{(s,a)}(\mu)
=
\frac{2\alpha_b}{1-\gamma}
\Bigl(
\sqrt{\Phi(\mathcal Y^{[b]};\mu)}
-
\sqrt{\Phi(\mathcal Y^{[b]};\mu)-\Delta(s,a;\mathcal Y^{[b]},\mu)}
\Bigr).
\end{equation}
Consequently, \(U_b(\mu)-U_b^{(s,a)}(\mu)\) is strictly increasing in \(\Delta(s,a;\mathcal Y^{[b]},\mu)\).
\end{lemma}

\begin{proof}
By the definition of \(\Delta\) in equation~\eqref{eq:def_delta},
\(
\Phi(\mathcal Y^{[b]}\cup\{(s,a)\};\mu)=\Phi(\mathcal Y^{[b]};\mu)-\Delta(s,a;\mathcal Y^{[b]},\mu).
\)
Substituting into \eqref{eq:def_bound_objective} for \(U_b\) and into the analogous expression for \(U_b^{(s,a)}\) yields \eqref{eq:bound_decrease_delta}. Applying the elementary identity \(a-b=(a^2-b^2)/(a+b)\) with \(a=\sqrt{\Phi(\mathcal Y^{[b]};\mu)}\) and \(b=\sqrt{\Phi(\mathcal Y^{[b]};\mu)-\Delta(s,a;\mathcal Y^{[b]},\mu)}\) yields the equivalent ratio form
\[
U_b(\mu)-U_b^{(s,a)}(\mu)
=
\frac{2\alpha_b}{1-\gamma}\,
\frac{\Delta(s,a;\mathcal Y^{[b]},\mu)}{\sqrt{\Phi(\mathcal Y^{[b]};\mu)}+\sqrt{\Phi(\mathcal Y^{[b]};\mu)-\Delta(s,a;\mathcal Y^{[b]},\mu)}}.
\]
For strict monotonicity, fix \(a:=\Phi(\mathcal Y^{[b]};\mu)>0\) and define \(f(x):=\sqrt{a}-\sqrt{a-x}\) for \(x\in[0,a]\). Since \(f'(x)=1/(2\sqrt{a-x})>0\), \(f\) is strictly increasing on \([0,a]\), and \(U_b(\mu)-U_b^{(s,a)}(\mu)\) is a positive constant multiple of \(f(\Delta(s,a;\mathcal Y^{[b]},\mu))\). Hence \(U_b(\mu)-U_b^{(s,a)}(\mu)\) is strictly increasing in \(\Delta(s,a;\mathcal Y^{[b]},\mu)\).
\end{proof}

\subsection{Normalization of the reachability measure}
\label{app:normalized_reachability}

Section~\ref{subsec:successor_guidance} obtains the successor-guided distribution \(\bar H_b\) by normalizing the positive part of the per-candidate scores \(H_b(s_i,a_i)\) of equation~\eqref{eq:reachability_score} over the unlabeled pool. We make the normalization explicit. Let \(\mathcal U_b:=\mathcal D\setminus\mathcal S_b\) denote the unlabeled set at round \(b\), and define
\[
Z_b
:=
\sum_{(s_j,a_j)\in\mathcal U_b}
[H_b(s_j,a_j)]_+,
\qquad
[x]_+:=\max\{x,0\}.
\]
The successor-guided distribution is
\[
\bar H_b(s_i,a_i)
:=
\begin{cases}
\dfrac{[H_b(s_i,a_i)]_+}{Z_b}, & (s_i,a_i)\in\mathcal U_b\text{ and }Z_b>0,\\[1.25ex]
\dfrac{1}{|\mathcal U_b|}, & (s_i,a_i)\in\mathcal U_b\text{ and }Z_b=0,\\[1.25ex]
0, & (s_i,a_i)\in\mathcal S_b.
\end{cases}
\]
Clipping to the positive part discards candidates whose successor overlap with the target set is negative under the current policy, so they receive no probability mass. The fallback to the uniform distribution on \(\mathcal U_b\) when \(Z_b=0\)---that is, when no unlabeled candidate has positive successor overlap with \(T_b\)---keeps \(\bar H_b\) a well-defined probability measure on \(\mathcal U_b\) at every round. Under this definition, the mixture
\(
\rho_b=\beta_b\hat d_{\mathrm{beh}}^{(b)}+(1-\beta_b)\bar H_b
\)
of equation~\eqref{eq:mixed_measure} is always a probability measure on the unlabeled pool, with the behavior component controlling coverage and the successor component concentrating on candidates that can reach \(T_b\) under \(\pi^{[b]}\).

\subsection{Exact-gain scoring complexity}
\label{app:complexity}

Section~\ref{subsec:successor_guidance} states that, at each round \(b\), evaluating the exact gain \(\Delta(s_i,a_i;\mathcal Y^{[b]},\rho_b)\) of equation~\eqref{eq:exact_gain} for every unlabeled candidate costs \(O(d^3+(|\mathcal D|-b)d^2)\) when the directional operator is formed explicitly. We record the per-round cost breakdown.

By Proposition~\ref{prop:exact_gain},
\[
\Delta(s_i,a_i;\mathcal Y^{[b]},\rho_b)
=
\frac{\bigl\|\Sigma(\rho_b)^{1/2}\Lambda_b^{-1}\phi(s_i,a_i)\bigr\|_2^2}{1+\sigma_b(s_i,a_i)^2}.
\]
A candidate-wise evaluation therefore needs:
\begin{enumerate}
\item the symmetric square root \(\Sigma(\rho_b)^{1/2}\), via one eigendecomposition of the symmetric \(d\times d\) matrix \(\Sigma(\rho_b)\), at \(O(d^3)\);
\item the matrix product \(A_b:=\Sigma(\rho_b)^{1/2}\Lambda_b^{-1}\in\mathbb R^{d\times d}\), at \(O(d^3)\);
\item one matrix--vector product \(A_b\phi(s_i,a_i)\) per unlabeled candidate, at \(O(d^2)\), followed by an \(O(d)\) squared-norm;
\item the scalar uncertainty \(\sigma_b(s_i,a_i)^2=\phi(s_i,a_i)^\top\Lambda_b^{-1}\phi(s_i,a_i)\), already maintained by the pessimistic reward pipeline.
\end{enumerate}
Summing over the \(|\mathcal D|-b\) unlabeled candidates, the per-round cost is \(O(d^3+(|\mathcal D|-b)d^2)\), as claimed. By contrast, the \SURE{} score of equation~\eqref{eq:swvr} requires only \(O((|\mathcal D|-b)d^2)\) per round to evaluate the \(\sigma_b(s_i,a_i)^2\) values across the unlabeled pool---the \(O(d^3)\) operator construction is avoided---while reusing the per-candidate scores \(H_b(s_i,a_i)\) and the mixture weights \(\rho_b(s_i,a_i)\) already produced by the surrogate-occupancy step.

\section{Theoretical Analysis}
\label{app:theoretical_analysis}

This appendix supports the discussion at the end of Section~\ref{subsec:successor_guidance}. We define a coverage coefficient that measures the cost of replacing the unknown optimal occupancy \(d^{\pi_r^\star}\) by a computable reference measure on the offline pool, prove the resulting unified suboptimality bound, derive a budget-dependent design rate that exposes the dependence on coverage mismatch and effective dimension, and explain how the \SURE{} score arises as a scalable surrogate for the exact one-step gain. Every formal statement is proved.

\subsection{Coverage transfer and unified bound}
\label{app:coverage_transfer}

\begin{definition}[Coverage coefficient and query complexity]
\label{def:cqry}
For any probability measure \(\nu\) on the unlabeled pool, define
\[
\kappa(\nu)
:=
\inf\bigl\{c\ge 1:\Sigma(d^{\pi_r^\star})\preceq c\,\Sigma(\nu)\bigr\},
\qquad
r_\nu
:=
\rank(\Sigma(\nu)),
\]
and the query-complexity coefficient \(\Cqry(\nu):=\kappa(\nu)\,r_\nu\). We call \(\nu\) \emph{admissible} if \(\kappa(\nu)<\infty\). Admissibility is a data-coverage condition on the offline pool: \(\kappa(\nu)<\infty\) requires \(\mathrm{range}(\Sigma(d^{\pi_r^\star}))\subseteq\mathrm{range}(\Sigma(\nu))\). If the pool leaves any feature direction used by the optimal policy uncovered, no reward-labeling strategy can recover it from labels alone.
\end{definition}

\begin{proposition}[Coverage transfer]
\label{prop:coverage_transfer}
For any admissible reference measure \(\nu\) and any labeled set \(\mathcal Y\),
\[
\Phi(\mathcal Y;d^{\pi_r^\star})
\le
\kappa(\nu)\,\Phi(\mathcal Y;\nu).
\]
\end{proposition}

\begin{proof}
By definition of \(\kappa(\nu)\), \(A:=\kappa(\nu)\Sigma(\nu)-\Sigma(d^{\pi_r^\star})\succeq 0\). Set \(M:=\Lambda(\mathcal Y)^{-1}\succ 0\). For any pair of positive semidefinite matrices we have \(\tr(AM)=\tr(M^{1/2}AM^{1/2})\ge 0\), since \(M^{1/2}AM^{1/2}\succeq 0\). Therefore
\[
0
\le
\tr\!\bigl([\kappa(\nu)\Sigma(\nu)-\Sigma(d^{\pi_r^\star})]M\bigr)
=
\kappa(\nu)\,\tr(\Sigma(\nu)M)-\tr(\Sigma(d^{\pi_r^\star})M).
\]
For any probability measure \(\mu\), trace cyclicity together with \(\sigma_b(s,a)^2=\phi(s,a)^\top\Lambda(\mathcal Y)^{-1}\phi(s,a)\) gives
\[
\tr(\Sigma(\mu)M)
=
\tr\!\bigl(\E_{(s,a)\sim\mu}[\phi(s,a)\phi(s,a)^\top]\Lambda(\mathcal Y)^{-1}\bigr)
=
\E_{(s,a)\sim\mu}\!\bigl[\phi(s,a)^\top\Lambda(\mathcal Y)^{-1}\phi(s,a)\bigr]
=
\Phi(\mathcal Y;\mu).
\]
Substituting \(\mu=\nu\) and \(\mu=d^{\pi_r^\star}\) yields \(\Phi(\mathcal Y;d^{\pi_r^\star})\le\kappa(\nu)\,\Phi(\mathcal Y;\nu)\).
\end{proof}

\begin{proposition}[Coverage preserved by the behavior mixture]
\label{prop:mixing}
Let \(\rho_b=\beta_b\hat d_{\mathrm{beh}}^{(b)}+(1-\beta_b)\bar H_b\) be the round-\(b\) reachability measure of equation~\eqref{eq:mixed_measure}. For any \(\beta_b>0\),
\[
\Sigma(\rho_b)\succeq\beta_b\,\Sigma(\hat d_{\mathrm{beh}}^{(b)}),
\qquad
\kappa(\rho_b)\le\frac{\kappa(\hat d_{\mathrm{beh}}^{(b)})}{\beta_b}.
\]
This shows that successor guidance can concentrate mass near currently high-value targets without discarding the feature directions already covered by the empirical behavior distribution.
\end{proposition}

\begin{proof}
By construction, \(\bar H_b\) is a probability measure on the unlabeled pool (Appendix~\ref{app:normalized_reachability}), so its second-moment matrix is
\[
\Sigma(\bar H_b)
=
\sum_i \bar H_b(s_i,a_i)\,\phi(s_i,a_i)\phi(s_i,a_i)^\top
\succeq 0
\]
as a conic combination of rank-one positive semidefinite matrices. Linearity of \(\Sigma(\cdot)\) under convex combinations gives
\[
\Sigma(\rho_b)
=
\beta_b\,\Sigma(\hat d_{\mathrm{beh}}^{(b)})
+
(1-\beta_b)\,\Sigma(\bar H_b)
\succeq
\beta_b\,\Sigma(\hat d_{\mathrm{beh}}^{(b)}),
\]
which is the first claim.

For the second claim, write \(c_{\mathrm{beh}}:=\kappa(\hat d_{\mathrm{beh}}^{(b)})\). By definition of \(\kappa\), \(\Sigma(d^{\pi_r^\star})\preceq c_{\mathrm{beh}}\,\Sigma(\hat d_{\mathrm{beh}}^{(b)})\). The first claim and \(\beta_b>0\) imply \(\Sigma(\hat d_{\mathrm{beh}}^{(b)})\preceq\beta_b^{-1}\Sigma(\rho_b)\). Composing the two PSD inequalities,
\[
\Sigma(d^{\pi_r^\star})
\preceq
c_{\mathrm{beh}}\,\Sigma(\hat d_{\mathrm{beh}}^{(b)})
\preceq
\frac{c_{\mathrm{beh}}}{\beta_b}\,\Sigma(\rho_b).
\]
By minimality in the definition of \(\kappa(\rho_b)\), \(\kappa(\rho_b)\le c_{\mathrm{beh}}/\beta_b=\kappa(\hat d_{\mathrm{beh}}^{(b)})/\beta_b\).
\end{proof}

Combining Theorem~\ref{thm:label_sensitive} with Proposition~\ref{prop:coverage_transfer} yields the structural bound under any admissible reference measure.

\begin{theorem}[Unified reference-measure suboptimality bound]
\label{thm:master}
Under Definition~\ref{def:linear_mdp}, equation~\eqref{eq:pessimistic_reward_gap}, and equation~\eqref{eq:offline_backbone_gap}, with probability at least \(1-2\delta\), for any admissible reference measure \(\nu\) and any labeled set \(\mathcal Y\) producing a pessimistic policy \(\hat\pi_{\tilde r}\) with confidence radius \(\alpha\),
\[
\SubOpt(\mathcal Q^{(B)})
\le
\Gamma_{\mathrm{off}}(\mathcal D,\delta)
+
\frac{2\alpha}{1-\gamma}
\sqrt{\kappa(\nu)\,\Phi(\mathcal Y;\nu)}.
\]
\end{theorem}

\begin{proof}
Theorem~\ref{thm:label_sensitive}, applied to \(\mathcal Y\) and the pessimistic policy \(\hat\pi_{\tilde r}\) with confidence radius \(\alpha\), gives
\[
\SubOpt(\mathcal Q^{(B)})
\le
\Gamma_{\mathrm{off}}(\mathcal D,\delta)
+
\frac{2\alpha}{1-\gamma}
\sqrt{\E_{(s,a)\sim d^{\pi_r^\star}}[\sigma(s,a)^2]},
\]
where \(\sigma\) is the uncertainty radius associated with \(\mathcal Y\). The trace-identity argument used in the proof of Proposition~\ref{prop:coverage_transfer} (taking \(\mu=d^{\pi_r^\star}\)) yields
\(
\E_{(s,a)\sim d^{\pi_r^\star}}[\sigma(s,a)^2]=\Phi(\mathcal Y;d^{\pi_r^\star}).
\)
Substituting and applying Proposition~\ref{prop:coverage_transfer},
\[
\Phi(\mathcal Y;d^{\pi_r^\star})
\le
\kappa(\nu)\,\Phi(\mathcal Y;\nu).
\]
The square-root map is monotone increasing on \([0,\infty)\), so substituting yields the displayed bound.
\end{proof}

\subsection{Budget-dependent rates}
\label{app:design_rate}

We translate Theorem~\ref{thm:master} into a budget-dependent rate by tracking how fast the design potential \(\Phi(\mathcal Y^{[B]};\nu)\) decays under exact-gain selection against a fixed reference measure.

\begin{theorem}[Fixed-measure greedy design rate]
\label{thm:fixed_measure_rate}
Fix an admissible probability measure \(\nu\) on the pool with rank \(r_\nu:=\rank(\Sigma(\nu))\), and run the exact-gain rule
\[
(s_{i_b},a_{i_b})\in\arg\max_{(s_i,a_i)\in\mathcal D\setminus\mathcal S_b}\Delta(s_i,a_i;\mathcal Y^{[b]},\nu)
\]
of Proposition~\ref{prop:exact_gain} against this fixed reference measure. Then for every \(B\ge 1\),
\[
\Phi(\mathcal Y^{[B]};\nu)
\le
\frac{2\,r_\nu}{B}.
\]
\end{theorem}

\begin{proof}
Write \(F_b:=\Phi(\mathcal Y^{[b]};\nu)\ge 0\). By Proposition~\ref{prop:exact_gain}, \(F_b\) is monotonically nonincreasing in \(b\) and the round-\(b\) drop equals the exact gain at the chosen candidate:
\[
F_b-F_{b+1}=\Delta(s_{i_b},a_{i_b};\mathcal Y^{[b]},\nu).
\]

\emph{Step 1: greedy lower bound.}
The rule maximizes \(\Delta(\cdot;\mathcal Y^{[b]},\nu)\) over the unlabeled pool, and \(\nu\) is supported on the pool, so
\[
\Delta(s_{i_b},a_{i_b};\mathcal Y^{[b]},\nu)
\ge
\E_{(s,a)\sim\nu}\!\bigl[\Delta(s,a;\mathcal Y^{[b]},\nu)\bigr].
\]

\emph{Step 2: lower bound on the expected gain.}
By Proposition~\ref{prop:exact_gain},
\[
\Delta(s,a;\mathcal Y^{[b]},\nu)
=
\frac{\|\Sigma(\nu)^{1/2}\Lambda_b^{-1}\phi(s,a)\|_2^2}{1+\sigma_b(s,a)^2}.
\]
Section~\ref{sec:prelim} takes \(\lambda\ge 1\) so that \(\Lambda_b\succeq I\) and \(\sigma_b(s,a)^2=\phi(s,a)^\top\Lambda_b^{-1}\phi(s,a)\le\|\phi(s,a)\|_2^2\le 1\). Hence the denominator is at most \(2\), and
\[
\E_{(s,a)\sim\nu}\!\bigl[\Delta(s,a;\mathcal Y^{[b]},\nu)\bigr]
\ge
\tfrac{1}{2}\,\E_{(s,a)\sim\nu}\!\bigl[\|\Sigma(\nu)^{1/2}\Lambda_b^{-1}\phi(s,a)\|_2^2\bigr].
\]
Expanding the squared norm,
\[
\|\Sigma(\nu)^{1/2}\Lambda_b^{-1}\phi(s,a)\|_2^2
=
\phi(s,a)^\top\Lambda_b^{-1}\Sigma(\nu)\Lambda_b^{-1}\phi(s,a)
=
\tr\!\bigl(\Lambda_b^{-1}\Sigma(\nu)\Lambda_b^{-1}\phi(s,a)\phi(s,a)^\top\bigr),
\]
using \(x^\top Mx=\tr(Mxx^\top)\). Taking expectation over \((s,a)\sim\nu\), with \(\E_{(s,a)\sim\nu}[\phi(s,a)\phi(s,a)^\top]=\Sigma(\nu)\) and trace cyclicity,
\[
\E_{(s,a)\sim\nu}\!\bigl[\|\Sigma(\nu)^{1/2}\Lambda_b^{-1}\phi(s,a)\|_2^2\bigr]
=
\tr\!\bigl(\Lambda_b^{-1}\Sigma(\nu)\Lambda_b^{-1}\Sigma(\nu)\bigr)
=
\tr(A_b^2)
=
\|A_b\|_F^2,
\]
where \(A_b:=\Sigma(\nu)^{1/2}\Lambda_b^{-1}\Sigma(\nu)^{1/2}\succeq 0\).

\emph{Step 3: relating \(\|A_b\|_F^2\) to \(F_b\).}
Trace cyclicity gives \(\tr(A_b)=\tr(\Lambda_b^{-1}\Sigma(\nu))=F_b\) (the last equality is the trace identity from Step~2). The matrix \(A_b\) has rank at most \(\rank(\Sigma(\nu))=r_\nu\), so its eigenvalues \(\lambda_1(A_b),\ldots,\lambda_{r_\nu}(A_b)\ge 0\) (with the remaining \(d-r_\nu\) eigenvalues equal to zero) satisfy, by the Cauchy--Schwarz inequality applied to the all-ones vector and \((\lambda_i)_{i\le r_\nu}\),
\[
F_b^2
=
\Bigl(\sum_{i=1}^{r_\nu}\lambda_i(A_b)\Bigr)^2
\le
r_\nu\sum_{i=1}^{r_\nu}\lambda_i(A_b)^2
=
r_\nu\,\|A_b\|_F^2.
\]
Therefore \(\|A_b\|_F^2\ge F_b^2/r_\nu\), and combining Steps~1--3,
\begin{equation}
\label{eq:fixed_rate_recursion}
F_b-F_{b+1}\ge\frac{F_b^2}{2\,r_\nu}.
\end{equation}

\emph{Step 4: telescoping the recursion.}
If \(F_b=0\) for some \(b\le B\), monotonicity gives \(F_B=0\) and the bound is trivial. Otherwise, dividing \eqref{eq:fixed_rate_recursion} by \(F_b F_{b+1}>0\) and using \(F_{b+1}\le F_b\),
\[
\frac{1}{F_{b+1}}-\frac{1}{F_b}
=
\frac{F_b-F_{b+1}}{F_b F_{b+1}}
\ge
\frac{F_b-F_{b+1}}{F_b^2}
\ge
\frac{1}{2\,r_\nu}.
\]
Summing over \(b=0,\ldots,B-1\) and using \(F_0\ge 0\) (so \(1/F_0\ge 0\)),
\[
\frac{1}{F_B}
\ge
\frac{1}{F_0}+\frac{B}{2\,r_\nu}
\ge
\frac{B}{2\,r_\nu},
\]
which gives \(F_B\le 2\,r_\nu/B\), the claimed bound.
\end{proof}

Combining Theorem~\ref{thm:master} with Theorem~\ref{thm:fixed_measure_rate} yields the budget-dependent corollary that we view as the headline rate statement: under exact-gain selection against a fixed reference measure, the label-dependent term decays at the parametric rate \(B^{-1/2}\) with a constant determined by the query complexity \(\Cqry(\nu)=\kappa(\nu)\,r_\nu\).

\begin{corollary}[Reference-measure suboptimality rate]
\label{cor:sgpds}
Under the assumptions of Theorem~\ref{thm:master} and exact-gain selection against an admissible fixed reference measure \(\nu\),
\[
\SubOpt(\mathcal Q^{(B)})
\le
\Gamma_{\mathrm{off}}(\mathcal D,\delta)
+
\frac{2\alpha_B}{1-\gamma}
\sqrt{\frac{2\,\Cqry(\nu)}{B}}.
\]
\end{corollary}

\begin{proof}
Theorem~\ref{thm:master} applied at the terminal round \(B\) with reference measure \(\nu\) and confidence radius \(\alpha=\alpha_B\) gives
\[
\SubOpt(\mathcal Q^{(B)})
\le
\Gamma_{\mathrm{off}}(\mathcal D,\delta)
+
\frac{2\alpha_B}{1-\gamma}
\sqrt{\kappa(\nu)\,\Phi(\mathcal Y^{[B]};\nu)}.
\]
By Theorem~\ref{thm:fixed_measure_rate}, \(\Phi(\mathcal Y^{[B]};\nu)\le 2\,r_\nu/B\), and the square-root map is monotone, so
\[
\sqrt{\kappa(\nu)\,\Phi(\mathcal Y^{[B]};\nu)}
\le
\sqrt{\kappa(\nu)\cdot\tfrac{2\,r_\nu}{B}}
=
\sqrt{\tfrac{2\,\Cqry(\nu)}{B}},
\]
using \(\Cqry(\nu)=\kappa(\nu)\,r_\nu\). Substituting yields the displayed bound.
\end{proof}

\subsection{Exact-gain structure and the \SURE{} score}
\label{app:swvr}

We finally explain why the \SURE{} score~\eqref{eq:swvr} approximates the exact gain~\eqref{eq:exact_gain}, and what is preserved when the directional operator \(\Sigma(\mu)^{1/2}\Lambda_b^{-1}\) is replaced by the scalar mixture weight \(\rho_b\).

For a fixed round \(b\), reference measure \(\mu\), and candidate \((s,a)\), define
\[
v_b(s,a):=\Lambda_b^{-1}\phi(s,a),
\qquad
\varrho_b(s,a)^2:=\|v_b(s,a)\|_2^2,
\qquad
u_b(s,a):=v_b(s,a)/\|v_b(s,a)\|_2\quad(\text{when } v_b(s,a)\neq 0).
\]
Define the directional relevance and the uncertainty-reduction factor
\[
\widetilde w_b(s,a;\mu):=u_b(s,a)^\top\Sigma(\mu)u_b(s,a),
\qquad
\zeta_b(s,a):=\frac{\varrho_b(s,a)^2}{1+\sigma_b(s,a)^2}.
\]
When \(v_b(s,a)=0\), set both quantities to zero.

\begin{proposition}[Exact-gain decomposition]
\label{prop:exact_decomposition}
For every round \(b\), every reference measure \(\mu\), and every candidate \((s,a)\),
\[
\Delta(s,a;\mathcal Y^{[b]},\mu)
=
\widetilde w_b(s,a;\mu)\,\zeta_b(s,a).
\]
\end{proposition}

\begin{proof}
By Proposition~\ref{prop:exact_gain},
\[
\Delta(s,a;\mathcal Y^{[b]},\mu)
=
\frac{\|\Sigma(\mu)^{1/2}\Lambda_b^{-1}\phi(s,a)\|_2^2}{1+\sigma_b(s,a)^2}.
\]
For any vector \(v\in\mathbb R^d\) and PSD \(\Sigma\), \(\|\Sigma^{1/2}v\|_2^2=v^\top\Sigma v\). Applying this with \(v=v_b(s,a)=\Lambda_b^{-1}\phi(s,a)\) and \(\Sigma=\Sigma(\mu)\),
\[
\|\Sigma(\mu)^{1/2}\Lambda_b^{-1}\phi(s,a)\|_2^2
=
v_b(s,a)^\top\Sigma(\mu)v_b(s,a).
\]
If \(v_b(s,a)=0\), the right-hand side is zero, so \(\Delta(s,a;\mathcal Y^{[b]},\mu)=0\); by definition \(\widetilde w_b(s,a;\mu)\,\zeta_b(s,a)=0\) as well, and the identity holds trivially. Otherwise \(v_b(s,a)=\varrho_b(s,a)\,u_b(s,a)\) and
\[
v_b(s,a)^\top\Sigma(\mu)v_b(s,a)
=
\varrho_b(s,a)^2\,u_b(s,a)^\top\Sigma(\mu)u_b(s,a)
=
\varrho_b(s,a)^2\,\widetilde w_b(s,a;\mu).
\]
Substituting,
\[
\Delta(s,a;\mathcal Y^{[b]},\mu)
=
\frac{\varrho_b(s,a)^2\,\widetilde w_b(s,a;\mu)}{1+\sigma_b(s,a)^2}
=
\widetilde w_b(s,a;\mu)\,\zeta_b(s,a),
\]
which is the claimed factorization.
\end{proof}

The \SURE{} score~\eqref{eq:swvr} replaces the two factors of Proposition~\ref{prop:exact_decomposition} by quantities already produced upstream. The directional relevance \(\widetilde w_b(\cdot;\mu)\) is replaced by the scalar mixture weight \(\rho_b(s,a)\)~--~which reuses the reachability scores of equation~\eqref{eq:reachability_score} and the behavior mixing of equation~\eqref{eq:mixed_measure}~--~and the uncertainty-reduction factor \(\zeta_b\) is replaced by the closely related \(\sigma_b(s,a)^2/(1+\sigma_b(s,a)^2)\), already maintained by the pessimistic reward pipeline. The next proposition records the cost saving that this substitution buys per round.

\begin{proposition}[Per-round cost of the \SURE{} score versus the exact gain]
\label{prop:swvr_complexity}
At each round \(b\), evaluating \(\SUREscore_b\) of equation~\eqref{eq:swvr} for every unlabeled candidate costs \(O((|\mathcal D|-b)\,d^2)\), whereas evaluating the exact gain \(\Delta(\,\cdot\,;\mathcal Y^{[b]},\rho_b)\) of equation~\eqref{eq:exact_gain} for every unlabeled candidate costs \(O(d^3+(|\mathcal D|-b)\,d^2)\) when the directional operator \(\Sigma(\rho_b)^{1/2}\Lambda_b^{-1}\) is formed explicitly.
\end{proposition}

\begin{proof}
The \SURE{} score requires only the per-candidate scalars \(\sigma_b(s_i,a_i)^2=\phi(s_i,a_i)^\top\Lambda_b^{-1}\phi(s_i,a_i)\) and the mixture weights \(\rho_b(s_i,a_i)\). Once \(\Lambda_b^{-1}\) is maintained incrementally by the pessimistic reward pipeline, each \(\sigma_b(s_i,a_i)^2\) costs one matrix--vector product and one inner product, i.e., \(O(d^2)\); summed over the \(|\mathcal D|-b\) unlabeled candidates this gives \(O((|\mathcal D|-b)\,d^2)\).

For the exact gain, Proposition~\ref{prop:exact_gain} requires the directional operator \(A_b:=\Sigma(\rho_b)^{1/2}\Lambda_b^{-1}\). Computing the symmetric square root \(\Sigma(\rho_b)^{1/2}\) needs one eigendecomposition of the symmetric \(d\times d\) matrix \(\Sigma(\rho_b)\), at \(O(d^3)\), and forming \(A_b\) needs one \(d\times d\) matrix product, also at \(O(d^3)\). Then for each unlabeled candidate, computing \(A_b\phi(s_i,a_i)\) costs \(O(d^2)\) and the squared norm an additional \(O(d)\). Summing the operator-construction cost and the \(|\mathcal D|-b\) per-candidate matrix--vector products gives \(O(d^3+(|\mathcal D|-b)\,d^2)\) per round, as claimed.
\end{proof}

\(\SURE{}\) thus preserves the relevance-times-uncertainty structure of the exact gain~--~each candidate is scored by the product of a relevance term and an uncertainty term~--~while avoiding the per-round \(O(d^3)\) operator construction. This is the structural reason why dropping either factor (\textsc{Reach} or \textsc{Uncertainty} in the empirical ablation of Section~\ref{sec:qualitative_ablation}) breaks the score whenever the dropped factor is not nearly flat across candidates.

\section{Experimental Details}
\label{app:experiments}

This appendix complements Section~\ref{sec:experiments} with domain and dataset details, baseline implementations, the shared partially labeled offline-RL pipeline, and acquisition hyperparameters.

\subsection{Domains and Datasets}
\label{app:domains_datasets}

We evaluate on five domains: the finite linear MDPs of Section~\ref{sec:linear_mdp}, constructed so that the assumptions of Section~\ref{sec:prelim} hold exactly, and four offline-RL domains chosen to span continuous navigation, continuous-control navigation at two scales, and high-dimensional locomotion. Within each domain, every reward-selection strategy and every random seed sees the same fixed reward-free transition pool \(\mathcal D=\{(s_i,a_i,s_i')\}_{i=1}^{|\mathcal D|}\); a strategy is allowed to acquire \(B\) transition-level reward labels from this pool.

\subsubsection{Domain visualizations}
\label{app:domain_viz}

\begin{figure}[t]
    \centering
    \includegraphics[width=\textwidth]{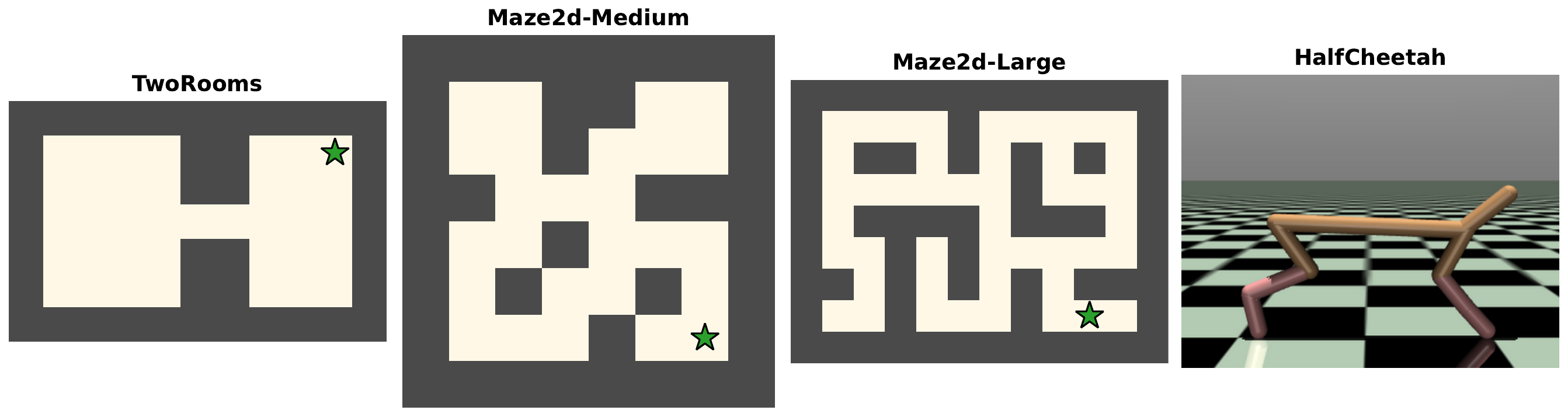} 
    \caption{Visualization of the four evaluation domains. From left to right: TwoRooms, Maze2d-Medium, Maze2d-Large, and HalfCheetah-Medium-Replay.}
    \label{fig:domain_overview}
\end{figure}

Figure~\ref{fig:domain_overview} shows one subplot per domain. The four subplots correspond to the following.

\begin{itemize}
\item \textbf{TwoRooms.} A continuous two-room gridworld in which two rectangular rooms are connected by a narrow doorway. The state is the agent's continuous cell and the action set is the four cardinal moves; a sparse reward is given only at the goal cell, which lies in the room opposite the start. The doorway concentrates the directions an optimal policy must traverse, providing a small-scale sanity check for the reward-selection problem.

\item \textbf{Maze2d-Medium.} A continuous 2D point-mass navigation domain from the D4RL benchmark suite~\citep{fu2020d4rl}. The state encodes the point-mass position and velocity, the action is a continuous 2D force, and a sparse reward is given upon reaching the goal. ``Medium'' refers to the layout size: a moderately long path between start and goal with a small number of corridors.

\item \textbf{Maze2d-Large.} The larger-layout counterpart of Maze2d-Medium, with the same continuous state and action spaces and the same sparse goal-reaching reward, but a longer required path between start and goal. The longer horizon concentrates the optimal-policy directions on a narrow corridor that occupies a smaller fraction of the offline pool, increasing the difficulty of credit assignment.

\item \textbf{HalfCheetah-Medium-Replay.} A continuous-control locomotion domain from the D4RL MuJoCo suite~\citep{brockman2016gym,fu2020d4rl}. The state encodes joint positions and velocities of the simulated half-cheetah (\(17\)-D), and the action is a continuous torque vector (\(6\)-D). The Medium-Replay dataset records the full replay buffer of a partially trained agent, so it mixes high- and low-quality transitions; together with the navigation domains it provides a higher-dimensional, non-navigation contrast for the acquisition comparison.
\end{itemize}

\subsubsection{Dataset collection}
\label{app:dataset_collection}

We use the same fixed reward-free transition pool \(\mathcal D\) for every reward-selection strategy and every random seed within a domain.

\paragraph{TwoRooms.} We collect the offline pool ourselves, with \(|\mathcal D|=15{,}000\) transitions. The data-collecting policy mixes uniform random exploration with goal-directed trajectories from a hand-coded shortest-path policy: at each step it takes a random action with probability \(0.5\) and a shortest-path action with probability \(0.5\). This mixture ensures coverage of both the corridor connecting the two rooms and the surrounding non-corridor cells.

\paragraph{Maze2d-Medium and Maze2d-Large.} We use the standard D4RL datasets ``maze2d-medium-v1'' (\(|\mathcal D|\approx 2\text{M}\)) and ``maze2d-large-v1'' (\(|\mathcal D|\approx 4\text{M}\)), both collected by a hand-engineered planner that drives the point mass to repeatedly resampled goals. We strip the rewards from the released datasets to obtain the reward-free pool used as input to reward selection.

\paragraph{HalfCheetah-Medium-Replay.} We use the D4RL dataset ``halfcheetah-medium-replay-v2'' (\(|\mathcal D|\approx 200\text{K}\)), which records the full replay buffer of a SAC agent trained to medium-level performance. Rewards are stripped to obtain the reward-free pool used as input to reward selection.

\subsection{Retained-Label Reward Selection Baselines}
\label{app:baselines}

All baselines share the offline pool \(\mathcal D\) and the same partially labeled offline-RL pipeline of Appendix~\ref{app:plorl_implementation}; they differ only in how each round's transition is selected from the unlabeled pool. We describe the implementation of each in turn.

\paragraph{Uniform.}
At round \(b\), Uniform draws a transition uniformly at random from the unlabeled subset \(\mathcal D\setminus\mathcal S_b\), without replacement across rounds. Uniform uses no value information, no reward-uncertainty information, and no policy estimate, and serves as the coverage-only baseline.

\paragraph{RLLF-Guided.}
RLLF-Guided is the guided heuristic of \citet{chaudhari2026data}. At round \(b\), with \(Q_b\) the action-value function of the current pessimistic policy \(\pi^{[b]}\) under reward \(\tilde r_b\) and \(\hat z^\star_b\in\arg\max_{(s,a,s')\in\mathcal D}Q_b(s,a)\) the current top-valued transition, each unlabeled candidate \((s_i,a_i)\) is scored by an empirical one-step transition probability of reaching \(\hat z^\star_b\) from \((s_i,a_i)\), estimated empirically from the offline transition pool. The score is then mixed with the empirical behavior visitation distribution as in \citet{chaudhari2026data}, and the candidate with the highest mixed score is selected. We re-run this entire procedure at every round.

\paragraph{IDRL.}
IDRL is the Information Directed Reward Learning method of \citet{lindner2021information}, originally proposed for online reward querying. At each round it selects the candidate label that is expected to be most informative for comparing the returns of candidate high-performing policies. To adapt IDRL to our offline setting, we replace its online rollout-based return estimation with FQE~\citep{le2019batch} on the fixed dataset \(\mathcal D\), so that policy-return estimates used to compute the information-gain criterion are derived from the same offline pool used by every other baseline. The candidate set, the posterior reward distribution, and the FQE evaluation are recomputed at every round.

\paragraph{Reach (single-factor ablation).}
Reach is a single-factor ablation of \SURE{} that drops the reward-uncertainty term. At round \(b\), it selects the candidate \((s_i,a_i)\in\mathcal D\setminus\mathcal S_b\) that maximizes the relevance weight \(\rho_b(s_i,a_i)\) from equation~\eqref{eq:mixed_measure}. Reach uses the same successor-feature reachability and behavior-mixing pipeline as \SURE{}, but ignores how much each label would shrink the reward-confidence radius.

\paragraph{Uncertainty (single-factor ablation).}
Uncertainty is the complementary single-factor ablation: it drops the relevance weight and selects the candidate that maximizes \(\sigma_b(s_i,a_i)^2/(1+\sigma_b(s_i,a_i)^2)\), using the current pessimistic-reward design matrix \(\Lambda_b\). Uncertainty uses no policy or successor information; it queries wherever the reward model is currently most uncertain.

\subsection{Partially Labeled Offline RL Implementation}
\label{app:plorl_implementation}

After a reward-selection strategy produces the retained labeled set \(\mathcal Y^{[b]}\), every method follows the same pipeline from Section~\ref{sec:prelim}: the retained labels are used to construct a conservative reward estimate, the reward-free transitions in \(\mathcal D\) are annotated with this estimate, and IQL is trained on the resulting dataset. This shared pipeline ensures that empirical differences come from the reward-selection rule rather than the offline-RL learner. All methods use the same representation and offline-RL implementation.

The IQL learner~\citep{kostrikov2021iql} hyperparameters are listed in Table~\ref{tab:hp_iql}; we follow standard IQL settings throughout.

\begin{table}[h]
\centering
\caption{IQL hyperparameters.}
\label{tab:hp_iql}
\begin{tabularx}{\linewidth}{lX}
\toprule
Hyperparameter & Value \\
\midrule
Hidden sizes for \(V\), \(Q\), \(\pi\) & \(256\times 256\) MLP, ReLU \\
Optimizer & Adam \\
Learning rate & \(3\times 10^{-4}\) (all three networks) \\
Batch size & \(256\) (D4RL); \(64\) (TwoRooms) \\
Discount \(\gamma\) & \(0.99\) (HC-MR); \(0.995\) (Maze2d-M, Maze2d-L); \(0.95\) (TwoRooms) \\
Target-network Polyak rate \(\tau_{\mathrm{soft}}\) & \(5\times 10^{-3}\) \\
Expectile parameter \(\tau\) & \(0.7\) (D4RL); \(0.5\) (TwoRooms) \\
Advantage-weighting temperature \(\beta\) & \(3.0\) (Maze2d-M, HC-MR); \(10.0\) (Maze2d-L); \(1.0\) (TwoRooms) \\
Reward scaling & fixed across methods \\
\bottomrule
\end{tabularx}
\end{table}

\subsection{Hyperparameters}
\label{app:hyperparams}

\paragraph{\SURE{} acquisition score.}
The acquisition-score hyperparameters are listed in Table~\ref{tab:hp_sure}. We set the top-\(K\) target-set size \(|T_b|\) by domain scale, using a smaller \(K\) for TwoRooms and a larger \(K\) for the continuous domains. We use a small constant behavior-mixing weight \(\beta_b=0.1\) in all experiments; this prevents the relevance weight from concentrating on isolated outliers under low-coverage early rounds.

\begin{table}[h]
\centering
\caption{\SURE{} acquisition-score hyperparameters.}
\label{tab:hp_sure}
\begin{tabularx}{\linewidth}{lX}
\toprule
Hyperparameter & Value \\
\midrule
Top-\(K\) target-set size \(|T_b|\) & \(5\) (TwoRooms); \(32\) (Maze2d-M); \(64\) (Maze2d-L); \(64\) (HC-MR) \\
Behavior-mixing weight \(\beta_b\) & \(0.1\) (constant across rounds) \\
Successor-feature ridge regularizer \(\lambda_{\mathrm{SF}}\) & \(10^{-3}\) \\
Successor-feature estimator & closed-form ridge (Appendix~\ref{app:m_pi}) \\
Warm-start labels \(B_0\) & \(64\), uniformly sampled; shared across methods and counted toward the budget \\
Acquisition batch size \(M\) & \(64\) \\
Refit cadence & every acquisition round (after each batch of \(M\) labels) \\
\bottomrule
\end{tabularx}
\end{table}

\subsection{Evaluation Protocol}
\label{app:evaluation}

We report mean return over \(50\) random seeds; shaded regions in Figure~\ref{fig:main_comparison} show standard error. Policies are evaluated with \(100\) rollouts per seed for the D4RL domains and \(200\) rollouts for TwoRooms. All methods are evaluated under the same budget grid and the same full-feedback reference. Due to hardware constraints, we run our jobs on CPUs.

\subsection{Computational Cost}
\label{app:wallclock}

The methods differ in the computation required for acquisition. \textsc{Uniform} selects labels without fitting an acquisition model and is therefore substantially cheaper. \textsc{Uncertainty} updates only the reward estimator and its uncertainty quantities, whereas \textsc{RLLF-Guided}, \textsc{Reach}, IDRL, and \SURE{} additionally update policy-, value-, successor-, or policy-evaluation-related quantities. \SURE{}'s method-specific overhead consists of recomputing successor features, constructing the high-value target set, evaluating reachability, and scoring the unlabeled candidates. As described in Appendix~\ref{app:algorithm}, these components are updated once per batch of \(M=64\) newly acquired labels rather than after every individual label.

We compare the wall-clock time of all methods under identical batch sizes, feedback budgets, hardware, and evaluation protocols. Across the evaluated domains, \SURE{} takes on average approximately \(1.05\times\) as long as \textsc{RLLF-Guided} and IDRL, and has comparable runtime to \textsc{Reach} and \textsc{Uncertainty}. The successor-feature and reachability computations used by \SURE{} and \textsc{Reach}, as well as the reward-uncertainty updates used by \SURE{} and \textsc{Uncertainty}, contribute only a small fraction of the total runtime relative to the repeated offline-learning updates. \textsc{Uniform}, which does not require adaptive model-based acquisition, remains the least computationally expensive method overall.

Relative to \textsc{Uniform}, the results expose an explicit label--compute trade-off. For example, on Maze2d-Large, \SURE{} uses approximately \(20\times\) more computation but reaches \(90\%\) of the matched full-feedback return with approximately \(30\times\) fewer reward labels. Thus, \SURE{} is more computationally expensive than simple non-adaptive selection, but its wall-clock cost is comparable to existing informed adaptive approaches while yielding substantially greater label efficiency.

\subsection{Sensitivity to the Warm-Start Size}
\label{app:warmstart}

Early in acquisition, the reward model and the reachability computed from its policy and value estimates may be unreliable. In all continuous-control experiments, adaptive acquisition therefore begins from \(B_0=64\) uniformly sampled reward labels (Appendix~\ref{app:algorithm}). To evaluate sensitivity to this choice, we vary the warm-start size \(B_0\in\{64,200,400\}\) while keeping the subsequent acquisition batch size fixed at \(M=64\). Table~\ref{tab:warmstart} reports the percentage feedback required to reach \(80\%\) of the matched full-feedback return, averaged over \(50\) paired seeds.

\begin{table}[h]
\centering
\caption{Percentage feedback required by \SURE{} to reach \(80\%\) of the matched full-feedback return under different warm-start sizes \(B_0\), averaged over \(50\) paired seeds.}
\label{tab:warmstart}
\begin{tabular}{lccc}
\toprule
Domain & \(B_0=64\) & \(B_0=200\) & \(B_0=400\) \\
\midrule
Maze2d-Large & \textbf{0.060\%} & 0.064\% & 0.067\% \\
HalfCheetah & \textbf{0.300\%} & 0.330\% & 0.360\% \\
\bottomrule
\end{tabular}
\end{table}

The feedback required by \SURE{} changes only marginally across the tested warm-start sizes, showing that its label efficiency is stable with respect to the initialization. This robustness is consistent with the design of the \SURE{} score: the reachability measure~\eqref{eq:mixed_measure} mixes the successor-guided distribution with the empirical behavior distribution through \(\beta_b\), so inaccurate early value estimates cannot fully determine the acquisition distribution, and the multiplicative uncertainty factor in~\eqref{eq:swvr} further requires selected transitions to be informative for the reward model. The mild advantage of \(B_0=64\) is also expected: fewer labels are spent on uniform initialization, allowing adaptive acquisition to begin earlier.

\section{Limitations}
\label{app:limitations}

Our theoretical analysis is stated under the linear-MDP model of Definition~\ref{def:linear_mdp} and presumes that the chosen reference measure is admissible (Definition~\ref{def:cqry}); when the offline pool fails to cover feature directions used by the optimal policy, no reward-labeling strategy can recover them.  Finally, the offline-RL backbones considered in this work output continuous actions only; extending retained-label reward selection to more challenging settings, such as parameterized action spaces that combine discrete and continuous components, could be a direction for future work~\citep{zhang2024model,zhang2026noise,zhang2026evidence,zhang2026latent}.

\end{document}